\pdfoutput=1

\documentclass[11pt]{article}
\usepackage[final]{acl}

\usepackage{algorithm} 
\usepackage{algpseudocode} 
\usepackage{times}
\usepackage{latexsym}

\usepackage[T1]{fontenc}

\usepackage[utf8]{inputenc}

\usepackage{microtype}

\usepackage{inconsolata}

\usepackage{graphicx}
\usepackage{subcaption}

\usepackage{bytefield}
\usepackage{tabularx}

\usepackage{todonotes}

\newcommand{\amRemove}[1]
{\textcolor{red}{}}

\newcommand{\rtRemove}[1]
{\textcolor{red}{}}

\title{A Novel Computational Modeling Foundation \\for Automatic Coherence Assessment}

\author{Aviya Maimon \and Reut Tsarfaty \\
        \texttt{aviyamn@gmail.com} \\ \texttt{reut.tsarfaty@biu.ac.il} \\
        Department of Computer Science, Bar Ilan University}
   
\begin{document}
\maketitle
\begin{abstract}
Coherence is an essential property of well-written texts, that refers to the way textual units relate to one another. 
In the era of generative AI, coherence assessment is essential for many NLP tasks; summarization, long-form question-answering, etc. 
Current NLP approaches for modeling coherence often rely on a proxy task, specifically sentence reordering. However, such an approach may not capture the full range of factors contributing to coherence.
To bridge this gap, in this work we employ the formal linguistic definition of \citet{Reinhart:1980} of what makes a discourse coherent, consisting of three conditions --- {\em cohesion, consistency} and {\em relevance} -- and formalize these conditions as respective computational tasks. 
We hypothesize that (i) a model trained on all of these tasks will learn the features required for coherence detection, and that (ii) a joint model for all tasks will exceed the performance of models trained on each task individually.
We evaluate this modeling approach on two human-rated coherence benchmarks: one of automatically-generated stories and one of real-world texts.
Our experiments confirm that jointly training on the proposed tasks leads to better performance on each task compared with task-specific models, and to better performance on assessing coherence overall, compared with strong baselines. 
Our formal coherence framework paves the way for advanced, broad-coverage automatic  assessment.

\end{abstract}

\section{Introduction}

The term {\em coherence} refers to the quality of texts where sentences and paragraphs flow smoothly and are logically connected, creating a clear and understandable progression of ideas.
Coherence detection is crucial for NLP tasks involving text quality measurements such as essay scoring or text quality assessment \cite{Somasundaran:2014, Feng:2014, Lai:2018}.
Its importance is further amplified nowadays in the era of large language models (LLMs). Ensuring coherent LLM outputs is essential for producing meaningful and understandable text for  generative tasks as summarization and question answering \cite{guan2021long, xu2018skeletonbased, yi2019coherent}, to name a few. 

The elusive 
nature of coherence makes it challenging for NLP systems to assess it effectively. While various linguistic theories of coherence exist (e.g., \citet{Halliday:1976, Joshi:1981, Givon:1995, Hobbs:1979, Dijk:1979, Mann:1988}), current approaches often rely on proxy tasks like \emph{sentence reordering} \cite{Lapata:2003, Miltsakaki:2000}. However, this    approach 
oversimplifies 
coherence, potentially leading to models that  struggle with real-world texts \cite{laban2021transformer}.
Furthermore, 
coherence is multifaceted, varying across genres, contexts, and styles. Proxy tasks often fail to capture this complexity, leading to models 
unable to generalize  across different domains and real-world settings.
On top of that, existing NLP models for coherence, such as \citet{Barzilay:2008}'s sentence reordering, often lack direct evaluation and are instead assessed on downstream tasks like readability \cite{guinaudeau-strube-2013-graph, mesgar-strube-2016-lexical, mesgar-strube-2018-neural} or essay scoring \cite{mesgar-strube-2018-neural, Somasundaran:2014, tay2017skipflow}. This approach can be expensive, biased towards the downstream task, and may overlook core aspects of coherence.

This work aims to provide a computationally workable definition of coherence, 
leveraging \citet{Reinhart:1980}'s theory which defines three conditions: \emph{cohesion, consistency}, and \emph{relevance}. 
Concretely, we propose to instantiate these conditions as computational  tasks, and  train them jointly to create a model that captures these properties.
We hypothesize that (i) such a model will effectively assess coherence, and (ii) shared information between tasks will improve task-specific performance.

To test our hypotheses, we implement a model trained jointly on tasks capturing \citeauthor{Reinhart:1980}'s conditions of coherence. The unified model incorporates five tasks
: sentence reordering \cite{Lapata:2003}, discourse relation detection \cite{miltsakaki-etal-2004-penn}, natural language inference \cite{inproceedings}, NP enrichment \cite{elazar2022textbased}, and irrelevant-sentence detection. 
Despite its relatively simple architecture, our model achieves SOTA results on most of these tasks.
We then evaluate the model's coherence assessment capability on two human-annotated benchmarks: the \emph{Grammarly Corpus of Discourse Coherence (GCDC)} \cite{Lai:2018} for real-world texts across four domains, and \emph{CoheSentia} \cite{anonymous:underreview} for artificially-generated stories. These benchmarks cover both natural and artificially-generated texts, human-rated for their levels of coherence.

Our empirical findings confirm our hypotheses, showing significant accuracy improvements and producing new SOTA results on both benchmarks. The joint model outperforms standalone models for individual tasks.
Our significant performance improvements demonstrate the model's efficacy in automatic coherence assessment. This framework paves the way for future models to not only identify incoherence, but also analyze its causes. Furthermore, integrating this methodology into text generation may lead to higher-quality outputs.

\rtRemove{
In sum, our work significantly advances NLP in several key aspects. 
We propose a refined coherence definition, fostering more accurate modeling techniques. 
We further establish a formal connection between coherence and a set of well-established NLP tasks, presenting a unified model that achieves SOTA performance on most of them. This unified model further demonstrates its capability by introducing a novel task, "irrelevant sentence recognition", which captures the crucial role of relevance in determining coherence. 
Finally, we showcase the model's broad applicability across diverse coherence benchmarks, encompassing varied domains, genres, and lengths.}
\rtRemove{
The remainder of this paper is organized as follows. In Section~\ref{Sec:Proposal}, we provide an overview of the linguistic foundation of our proposal. Later on in Section~\ref{section:coherenceTasks} we use those conditions as our coherence detection engine, where we map those conditions into a set of NLP tasks. Subsequently, in Section~\ref{Sec:Exp}, we delve into the details of our implementation of the coherence model using proxy tasks. We then provide the experimental setup for evaluating the coherence model in Section~\ref{Sec:CoherenceEvaluation}. Sections~\ref{Sec:ResultsTask} and ~\ref{Sec:ResultsCoherence} are dedicated to presenting the results of our experiments, for both the standalone proxy tasks as well as the unified model, and for evaluating the resulting model on the coherence scoring tasks. In Section~\ref{Sec:RelatedWork}, we highlight the existing body of related work, and lastly, in Section~\ref{Sec:Conclusion} we provide concluding remarks and directions for future endeavors.
}
\section{The Proposal: Coherence \`{a} la Reinhart} \label{Sec:Proposal}
This work aims to provide a computationally  workable
definition of coherence  by adopting \citet{Reinhart:1980}'s formalization, which identifies coherence through three criteria: \emph{Cohesion}, \emph{Consistency}, and \emph{Relevance}.
According to \citeauthor{Reinhart:1980}, a text is coherent only if it meets all three criteria. 
Recently, \citet{anonymous:underreview} used this framework to create a benchmark for coherence scoring of GPT-generated text, with human scores for these criteria. Here we take a different approach, using these conditions as the basis for computational modeling, aiming to predict the ingredients of these three properties via jointly trained tasks.

\paragraph{Cohesion}
The cohesion condition focuses on the formal elements that link sentences together.\footnote{The terms ``coherence'' and ``cohesion'' may be confusing to non-linguists. Cohesion relates to the surface forms used (e.g., connectors, pronouns), while coherence pertains to the overall semantics and flow of ideas.}
\citeauthor{Reinhart:1980} states that a text is cohesive if, for every sentence pair, at least one of two  conditions is met:

\emph{(1) Referentially linked:} 
A pair of sentences $\langle S_1,S_2 \rangle$ is referentially linked when $S_2$ references an entity mentioned in $S_1$.
A simple  example is using a pronominal anaphor: 
\begin{quote}
    ``\underline{Dan} is nice. Even Su likes \underline{him}.''
\end{quote}
 Here, the underlined entities co-refer. Other types of referential links are prepositional  links \cite{elazar2022textbased} or bridging anaphora \cite{hou-2021-end-end}.

\emph{(2) Linked by a semantic sentence connector}. 
A pair of sentences $\langle S_1,S_2 \rangle$ is connected if a discourse relation links them. 
These connectors indicate semantic relations like cause and effect, comparison, contrast, and more \cite{prasad-etal-2008-penn}.
An example of  linking by a semantic connector is:
\begin{quote}
    ``It was raining. \underline{So}, we stayed inside.''
\end{quote}
The sentences are cohesive due to the existence of the ``So'' connector. These connectors may be explicit or implicit \cite{pitler-etal-2009-automatic}. 


\paragraph{Consistency}

The consistency condition pertains to the formal semantic aspects of a text, ensuring \emph{logical} coherence, which is crucial for interpreting and deriving meaning.
Formally, this condition requires that for a set of sentences $\{S_i\}_{i=0}^{n-1}$, the meaning of each sentence $S_{i}$  must be consistent with all previous sentences $\{S_j\}_{i=0}^{i-1}$. This means all sentences can be true within a single world, not violating this worlds  assumptions and restrictions. An example of a violation is shown below:
\begin{quote}
``My father is dead now. That’s why 
he has decided to smoke a pipe'' \cite{freeman1966perseveration}
\end{quote}
Despite being cohesive (anaphora \& connectors), the passage lacks coherence due to world knowledge violations (a deceased cannot decide).\footnote{The consistency condition was further explored by \citet{honovich2021q2} to enhance the reliability of automatically generated texts.}

\paragraph{Relevance}

The relevance condition involves \emph{pragmatic} aspects, imposing constraints on the relationships of all sentences $\{S_i\}_{i=0}^{N-1}$ to the discourse topic and other contextual elements.
An example of a violation of this condition is as follows: 
\begin{quote}
    ``I poured some chemical into a beaker. 
    The chemical fell on my hand. 
    The professor immediately took me to the emergency bath. He is a great musician.''
\end{quote}

The last sentence is cohesive and consistent with the previous sentences but is irrelevant to the overall context and topic of the story.


All in all, Reinhart's theory outlines conditions encapsulating the fundamental aspects of coherence to determine text coherence. We propose designing NLP tasks to detect these properties.


\section{Research Hypotheses and Tasks}\label{section:coherenceTasks}
At the core of our approach is the implementation of Reinhart's coherence conditions as computational tasks, using a {minimal} set of NLP tasks designed to capture the features of cohesion, consistency, and relevance.
We hypothesize that a model trained jointly on {\em all} tasks will detect coherence effectively, and will outperform models trained on each task individually.


To verify this, we define five tasks reflecting these coherence conditions. 
\paragraph
{\bf The Sentence Reordering (SRO) Task}
This self-supervised task, proposed by \citet{Lapata:2003}, involves reordering shuffled sentences to restore their original coherent form.
%
%
%
For example, given the following input: 
``(1) Finally, the parser is evaluated. (2) We develop a useful parser. (3) Then we present our parser. (4) We first describe the older one.'' 
the correct order is $(2) \rightarrow (4) \rightarrow (3) \rightarrow (1)$.

Extending prior work on natural sentence order and coherence \cite{Lin:2011}, a model excelling at paragraph reconstruction should capture syntactic and semantic relationships between sentences, reflecting both {\em cohesion} and {\em consistency}.

\paragraph{The Discourse-Relation Recognition (DRR) Task}

Given a pair of sentences (discourse units - DUs), we aim to predict their discourse relation, reflecting notions such as cause and effect, comparison, and contrast.
%
%
%
%
For example, with the following input:
``John worked all night. He slept all day today.''
the model is expected to detect a relation marker reflecting \emph{contingency} (e.g., `so', `hence').

The discourse relation identification task enhances the model's ability to connect sentences, addressing the second sub-condition of \emph{cohesion}. 
\paragraph{The NP Enrichment (NPE) Task}
Introduced by \citet{elazar2022textbased}, the NPE task identifies prepositional links between noun phrase (NP) entities. Given NP pairs 
, it determines the existence of a prepositional relation and identifies the best preposition 
describing it $p(NP_1 NP_2)$. 
For a paragraph with \(k\) NP entities, the model outputs the prepositional links for all NP pairs where such a relation exists (or NONE otherwise).
%



For example, in the paragraph:
``Crown Princess Mary of Denmark gives birth to a male child.''
there are 4 NPs and thus 12 NP pairs. 
Sample outputs for these NP pairs are: 
(1)~in(birth, Denmark) and (2)~of(birth, male child).

A model trained on this task captures referential links between different parts of the discourse, serving as a proxy for the referential linking sub-condition of \emph{cohesion}.

\paragraph{The Natural Language Inference (NLI) Task}
The NLI task \cite{bowman2015large} aims to determine the truth value of a hypothesis based on a given premise. 
%
%
%
For example, given the premise: ``John inspects the uniform of a figure in some East Asian country.'' and the hypothesis: ``John is sleeping.'' the output will be a {\em contradiction}.

NLI evaluates NLP models' ability to capture logical relationships between sentences, serving as a proxy for the \emph{consistency} condition.

\paragraph{The Irrelevant Sentence Recognition (ISR) Task}
We propose a self-supervised task where the model identifies irrelevant sentences in a coherent paragraph. Given a paragraph with $N$ sentences, including one irrelevant sentence $s$, the model detects and outputs the irrelevant sentence.



For example, given the following input:
``(1) Rick is a helpful kid.
(2)~He does the dishes.
(3)~He avoids doing his homework.
(4)~He helps older people.'' 
The irrelevant sentence is (3).
    
The model is trained to assess sentence relevance to the overall topic and context, acting as a proxy for the \emph{relevance} condition.

\paragraph{Putting It All Together:}
%
We propose a Multi-Task Learning (MTL) approach, where a model is jointly trained on these tasks to capture all coherence conditions outlined by \citeauthor{Reinhart:1980}. 
This method leverages shared information during training, with the goal of enhancing both overall coherence detection and individual task performance. To assess coherence, we define two types of tasks:



\begin{itemize}
    \item {\bf The Coherence Scoring Task}
To confirm our hypothesis that the proposed model captures coherence, we evaluate its performance on the coherence scoring task, where given a paragraph $P$ the model predicts the coherence score $C$ as a human reader would. 



\item {\bf The Coherence Reasoning Task}
To examine conditions contributing to incoherence (cohesion, consistency, relevance) beyond a final score, we use the coherence reasoning task proposed by \citet{anonymous:underreview}. 
Given a paragraph $P$ and a new sentence $s$, the model predicts whether $s$ is \emph{cohesive}, \emph{consistent}, or \emph{relevant} to $P$ using distinct classifiers. 



\end{itemize}
We hypothesize that utilizing the MTL-powered architecture will improve the results on both coherence scoring and coherence reasoning.

\section{Coherence Assessment Setup} \label{Sec:CoherenceEvaluation}
Here, we detail the models and experimental setup for the coherence assessment tasks we define.

\subsection{The Coherence Scoring Task}
\begin{table*}[t]
    \centering
    \scalebox{0.8}{
    \begin{tabular}{|m{5em}||c|c|c||c|c|c|c|}
        \hline
         Dataset            & \multicolumn{3}{c||}{Split}& \multicolumn{4}{c|}{Per Instance}   \\
                            & Train & Validation & Test  & Max \#tokens& Avg \#tokens & Max \#sent.\ & Avg \#sent.\ \\
         \hline
         \hline
         GCDC               & 3.6k  & 800        & 800   & 333     &  156     &  10        & 32  \\
         \hline
         CoheSentia         & 350   & 75         & 75    & 226     &  150     &  15       & 6.5 \\
         \hline
    \end{tabular}
    }
    \caption{Main Statistics on the Datasets for Coherence Scoring}
    \vspace{-10pt}
    \label{Tab:coherenceDatasetSize}
\end{table*}

\paragraph{Models:}
We use two architectures for coherence scoring: Classification-Based (BERT \cite{devlin-etal-2019-bert}) 
and Generation-Based (T5 \cite{raffel2020exploring}). The model predicts for  a given text 3-way or 5-way scores, depending on the dataset.

In the Classification-Based models, the coherence score $C$ is predicted given the text $P$ using a classification head. 

In Generation-Based models, the input includes the text with dataset-specific prompts and an output. Example prompts and outputs are in Appendix~\ref{App:T5Prompts}.

Further details on experimental settings are in Appendix~\ref{App:CoherenceExperimentalSettings}.

\paragraph{Datasets and Evaluation:} 
We evaluate our model on two datasets: GCDC \cite{Lai:2018} and CoheSentia \cite{anonymous:underreview}. GCDC includes real-world text from various domains (Clinton emails, Enron emails, Yahoo Answers, Yelp reviews) with coherence scores from 1 (not coherent) to 3 (highly coherent). CoheSentia features GPT-3 generated stories (fiction and non-fiction) with scores ranging from 1 to 5. We use the ``incremental final score'' for CoheSentia stories. Dataset sizes and splits are detailed in Table~\ref{Tab:coherenceDatasetSize}.


To remain compatible with \citet{Lai:2018}, we use accuracy as the metric for evaluating the final coherence score of the text.






\paragraph{Baselines:}
To assess the effectiveness of our proposed model, we compare its performance on each dataset to the current SOTA models. 
For GCDC \citet{Lai:2018} introduced the ParSeq model with three stacked LSTMs for sentence, paragraph, and document embeddings, followed by a coherence classification layer. The SOTA model for GCDC by \citet{liu2023modeling} uses a multi-step approach: identifying document graph structures, converting subgraphs, constructing corpus-level graphs based on shared subgraphs, and encoding connections with a GCN.

For CoheSentia \citet{anonymous:underreview} created the SOTA model using a prompt-based approach with Flan-T5-large to assess coherence by adding a question at the beginning of each text. 

\subsection{The Coherence Reasoning Task}
\paragraph{Models:}
In Classification-Based models, the reasoning for each coherence attribute given the paragraph $P$ and the new sentence $s$ is predicted using a classification head. 

In Generation-Based models, the input includes the text with prompts and an output. Prompts and outputs for both datasets are in Appendix~\ref{App:T5Prompts}.

\paragraph{Datasets and Evaluation:}
We evaluate our model on the CoheSentia corpus \cite{anonymous:underreview}, which contains automatically generated stories with human annotations for cohesion, consistency, and relevance. We use precision, recall, and F1 scores for each property.

\paragraph{Baselines:}
We evaluate our model's effectiveness on the CoheSentia dataset by comparing it to the current SOTA model by \cite{anonymous:underreview}, which uses a prompt-based approach with Flan-T5-large, adding a question at the beginning of each text to assess coherence.

\section{Task-Specific Experimental Setup} \label{Sec:Exp}
In this section, we elaborate on the modeling  of the coherence proxy tasks. 
%
For each task, we evaluate two model variants: Classification-Based and Generation-Based, as detailed below.
Table~\ref{Tab:datasetSize} summarizes the datasets, evaluation metrics, and key statistics for each task 
(see further elaboration in Appendix~\ref{App:TasksExperimentalSettings}).
For the Generation-Based models, prompts and outputs  are detailed in Appendix~\ref{App:T5Prompts}.



\paragraph{The Sentence Reordering Models}

For the Classification-Based models, we adopt the topological sort architecture from \citet{Prabhumoye:2020}. 
Each paragraph's sentence pairs are represented as triplets $\langle S_i, C_k, S_j\rangle$, indicating whether $S_i$ precedes or follows $S_j$.  
The model has two stages: a binary classification head that predicts the pairwise relations and a second stage that produces the predicted order 
using the topological sort algorithm \cite{Tarjan:1976}. 

The Generation-Based models use prompts and predict the outputs with the final order.


\paragraph{The Discourse-Relation Recognition Models}

In the Classification-Based models, the input consists of a pair of DUs: $\langle DU_1, DU_2 \rangle$. A classification head predicts the discourse relation between them.
 
In the Generation-Based models, the input is an argument pair, and the model employs a chain-of-thought (CoT) method \cite{wei2023chainofthought} to predict the discourse relation.\footnote{CoT detection of discourse relations outperformed simpler prompts in our preliminary experiments.}
The CoT structure is $\langle$connector$\rangle \rightarrow \langle$$l_1$ relation$\rangle \rightarrow \langle$$l_2$ relation$\rangle$. 
That is, the model adopts a three-stage approach to predicting the $L_2$ relation type. 
The model first infers the implicit connective, then generalizes it to a broader relation category.

\paragraph{The NP Enrichment Models}


In the Classification-Based models, we extend the Bi-Affine architecture from \citet{dozat2017deep} to predict preposition relations between NP pairs instead of syntactic dependency labels. NP embeddings are created by pooling tokens representing each NP, and the model head predicts the preposition using the NP's anchor and complement representations (Figure~\ref{Fig:tokenHead} in Appendix).
%

The Generation-Based models predict prepositional relations for each NP pair 
independently. The input consists of the document text and a prompt specifying the NP pair.

%

\paragraph{The NL Inference Models}


Classification-Based models predict relations between the premise and hypothesis. 

Generation-Based models use prompts and predict outputs for premise-hypothesis pairs.

\paragraph{Irrelevant Sentence Recognition Models}

The Classification-Based models have two stages. 
Sentence pairs form triplets $\langle S_i, C_k, S_j\rangle$, where $C_k = 0|1$ indicates relevance. A binary classification head determines the relation.
In the second stage, the sentence with the lowest combined relations score is deemed irrelevant.
%

Generation-Based models use prompts and predict the irrelevant sentence as the output.

\paragraph{The Overall Joint Architecture}
To test our hypotheses, we implemented both Classification-Based and Generation-Based models, trained to solve all tasks jointly.

\begin{figure}[t]
\includegraphics[width=0.6\textwidth]{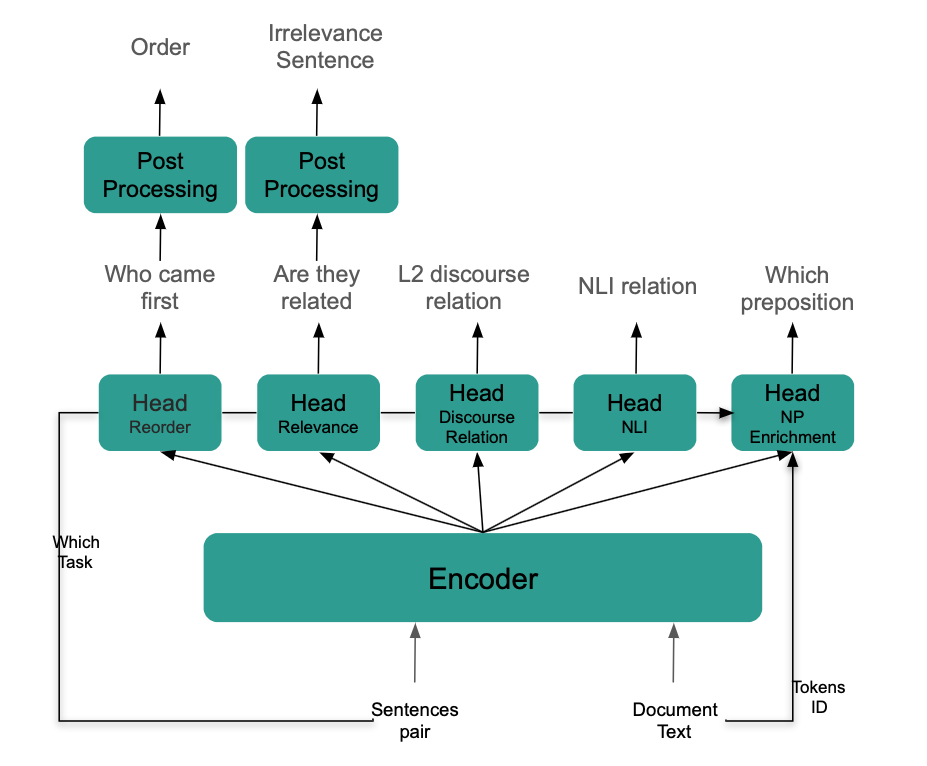}
\caption{Illustration of the encoder-only model where the input is a pair of sentences (most tasks) or for NPE task the input is a document and the token IDs of different NPs}
\vspace{-12pt}
\label{Fig:modelArch}
\end{figure}

For the Classification-Based variant, we use MTL \cite{caruana1997multitask} with a BERT encoder shared across tasks, each with a unique classification head. Each head predicts task-specific outputs (see Fig.~\ref{Fig:modelArch}). 
%
To address forgetting in MTL \cite{DBLP:journals/corr/GoodfellowMDCB13}, we implement an interleaved training strategy, alternating tasks in batches, ensuring each batch contains samples from a distinct task, effectively mitigating forgetting during MTL training.

For the Generation-Based model, we use the T5 encoder-decoder model, which allows concurrent fine-tuning of multiple tasks using distinct prompts. The prompt structure remains the same as individual task fine-tuning, but batches contain samples from specific tasks, distinct from the previous batch.


\begin{table*}[t]
    \centering
    \scalebox{0.75}{
    \begin{tabular}{|m{2em}||c||c||c|c|c||c|c|c|c|}
    \hline
         Task               & Dataset    & Metrics    & \multicolumn{3}{c||}{Split}& \multicolumn{4}{c|}{Per Instance}   \\
                                     &            &            & Train & Dev & Test  & Max     & Avg      &Max     & Avg \\
                                     &            &            &       &            &       & \#toks& \#toks &\#sent. & \#sent.\\
         \hline
         \hline
         SRO                & RocStories  & PMR {\fontsize{10}{10}\selectfont\cite{Chen:2016}}       & 68k   & 14k        & 14k   & 135     &  57      &  5        & 5  \\
                            & {\fontsize{10}{10}\selectfont\cite{mostafazadeh2016corpus}} & Acc {\fontsize{10}{10}\selectfont\cite{Logeswaran:2018}} &    &         &    &      &        &          &   \\
         \hline
         ISR                & RocStories {\fontsize{10}{10}\selectfont\cite{mostafazadeh2016corpus}} & Accuracy   & 68k   & 14k        & 14k   & 152     &  77      &  6        & 6  \\
         \hline
         DRR                & PDTB3 {\fontsize{10}{10}\selectfont\cite{AB2/SUU9CB_2019}}     & Accuracy   & 17.5k & 1.7k       & 1.5k  & 556     &  30      &  2        & 2  \\
         \hline
         NPE                & TNE {\fontsize{10}{10}\selectfont\cite{elazar2022textbased}}       & F1, Precision \& Recall & 3.5k  & 500        & 500   & 284     &  163     & 15        & 6.9\\
         \hline
         NLI                & MNLI {\fontsize{10}{10}\selectfont\cite{N18-1101}}      & Accuracy   & 393k  & 7.5k       & 2.5k  & (194,70)& (20,10)  & (8,8)     & (2,2) \\
         \hline
    \end{tabular}
    }
    \caption{The datasets and metrics used for each task and the train/dev/test split size with the max and average number of tokens and sentences. For the NLI task (x,y) refer to the numbers of (premise, hypothesis) respectively}
    \label{Tab:datasetSize}
\end{table*}

\section{Results} \label{Sec:ResultsTask}

We first aim to test the hypothesis that a model jointly trained on  tasks reflecting the different coherence conditions will effectively assess coherence. Table~\ref{Tab:CoherenceResults} shows the coherence scores of our jointly fine-tuned model (Ours-ALL) on the  GCDC and CoheSentia datasets, compared to current SOTA models on either dataset.
%
%
Compared to these models, our jointly fine-tuned model shows significant improvements in coherence scoring, especially on CoheSentia. We observe a 15\% and  27\%  accuracy gain for GCDC and  CoheSentia respectively, demonstrating that our proposed approach and selected tasks effectively contribute to capturing fundamental aspects of coherence.

We further analyze the contribution of the proxy tasks (Ours-All) by comparing it to a model without such fine-tuning (Ours-None) to isolate performance gains. As evident in Table~\ref{Tab:CoherenceResults} these tasks dramatically enhance performances. 
These results are supported by further qualitative analysis in Appendix~\ref{App:QualitativeExample}.


\begin{table}[t]
    \centering
    \scalebox{0.7}{
    \begin{tabular}{|m{15em}||c|c|}
    \hline
         Model                        & GCDC      & CoheSentia\\
         \hline
         \hline
         \citet{Lai:2018}             & 57.5      & ---       \\
         SOTA                         & 61.2     & 35.3      \\         
         \hline
         \hline
         Ours-None (bert-large)       & 50.2      & 34.3  \\
         Ours-ALL (bert-large)        & 72.5      & 55.7   \\
         \hline
         Ours-None (t5-large)         & 56.3      & 34.8 \\
         Ours-ALL (t5-large)          & \textbf{76.4} & \textbf{62.3} \\
         \hline
         \hline
         Controlled-nonCoherence (t5-large) & 52.8      &  36.8  \\
         \hline
         \hline
    \end{tabular}}
    \caption{Accuracy on Coherence Scoring
    The SOTA for GCDC is by \citet{liu2023modeling} and for CoheSentia is  \citet{anonymous:underreview}
    }
    \label{Tab:CoherenceResults}
    \vspace{-10pt}
\end{table}


Next we analyze the MTL model's success on assessing the coherence conditions (cohesion, consistency, relevance), by fine-tuning on the coherence reasoning task. We compare the results to SOTA from \citet{anonymous:underreview}, who fine-tuned the Flan-T5 model with a simple prompt, and to our model without initial fine-tuning on the coherence proxy tasks (Ours-None).
Table~\ref{Tab:coherenceReasoning} summarizes the coherence reasoning task results for all attributes and metrics. Our model achieves SOTA performance across all coherence conditions, demonstrating the efficacy of our approach.

\begin{table*}[t]
    \centering
    \scalebox{0.7}{
    \begin{tabular}{|m{12em}||c|c|c||c|c|c||c|c|c||}
    \hline
         Model                       & \multicolumn{3}{c||}{Cohesion}       & \multicolumn{3}{c||}{Consistency}  & \multicolumn{3}{c||}{Relevance}    \\
                                     & Precision  & Recall    & F1          & Precision& Recall     & F1         & Precision  & Recall    & F1        \\
         \hline
         \hline
         SOTA                        & 72.4       & 72.1      & 72.2        &  59.6    & 67.5       & 63.3       & 56.4       & 74.6      & 59.5      \\
         \hline
         \hline
         Ours-None (bert-large)      & 66.4       & 59.4      & 62.7        & 60.4      & 56.5      & 59.6       & 49.2       & 49.9      & 49.5      \\
         Ours-ALL (bert-large)       & 74.7       & 70.5      & 72.5        & 70.6      & 68.2      & 69.3       & 59.8       & 61.1      & 60.4      \\
         \hline
         Ours-None (t5-large)        & 81.1       & 80.3      & 80.7        & 60.4      & 62.6      & 61.5       & 48.1       & 49.6      & 48.8      \\
         Ours-ALL (t5-large)         & \textbf{83.1}&\textbf{83.2}& \textbf{83.1} &\textbf{78.5}&\textbf{80.3}&\textbf{79.4} &\textbf{70.8} &\textbf{76.9}&\textbf{73.7}\\
         \hline
    \end{tabular}}
    \caption{Results for Coherence Reasoning Task. The SOTA is by \citet{anonymous:underreview}}
    \label{Tab:coherenceReasoning}
\end{table*}



Finally, We evaluate the task-specific performance of models trained with either individual or joint fine-tuning on the proxy tasks, using both the Classification-Based and Generation-Based variations. Results are in Table~\ref{Tab:allResults}, alongside comparisons to current SOTA on these benchmarks.
Our findings consistently show that Generation-Based models outperform Classification-Based ones. More importantly, joint fine-tuning across all tasks consistently surpasses individual fine-tuning, particularly in the SRO, ISR, and DRR tasks, where it leads to significant performance improvements and even surpasses SOTA benchmarks. For the NPE task, joint fine-tuning achieves substantial recall gains, though precision falls short of SOTA results, offering a more balanced performance. The exception is the NLI task, where our model's performance is lower than SOTA.


\begin{table*}[t]
    \centering
    \scalebox{0.7}{
    \begin{tabular}{|m{12em}||c|c||c||c||c|c|c||c|}
    \hline
         Model                       & \multicolumn{2}{c||}{SRO} & ISR     & DRR       & \multicolumn{3}{c||}{NPE}  & NLI           \\
                                     & PMR  & ACC                & Accuracy& Accuracy  & F1   & P             & R   & Accuracy      \\
         \hline
         \hline
         SOTA                        & 81.9 & 90.8               &  -      & 64.7      & 64.0 & \textbf{80.5} & 53.1& \textbf{92.0} \\
         \hline
         \hline
         Ours-Individual (bert-large)& 51.8          & 69.5         & 60.4          & 60.0          & 53.1         & 67.1 & 44.0          & 87.4 \\
         Ours-ALL (bert-large)       & 67.1          & 83.2         & 78.6          & 65.7          & 64.4         & 79.8 & 54.2          & 90.2 \\
         \hline
         Ours-Individual (t5-large)  & 75.7          & 87.8         & 80.4          & 64.8          & 59.8         & 68.5 & 53.1          & 89.9 \\
         Ours-ALL (t5-large)         & \textbf{83.8} & \textbf{92.1}& \textbf{82.2} & \textbf{67.3} & \textbf{76.7}& 76.7 & \textbf{76.7} & 91.5\\
         \hline
    \end{tabular}}
    \caption{Results for all proxy tasks compared to SOTA performances. The SOTA model for SRO is ReBART \cite{basu-roy-chowdhury-etal-2021-everything}, for DRR is Contrastive Learning \cite{long2023facilitating}, for NPE is TNE \cite{elazar2022textbased} and for NLI T5-11B \cite{raffel2020exploring}}
    \label{Tab:allResults}
    \vspace{-10pt}
\end{table*}

In summary, our MTL model outperforms single-task models on all tasks, achieving SOTA results except for NLI, in line with our hypothesis on the benefits of the joint architecture.
Moreover, our MTL model, jointly trained on coherence proxy tasks, significantly improves performance, enhancing coherence scoring for both datasets and excelling in coherence reasoning, in line with the second part of our hypothesis.




\section{Analysis} 

\subsection{The Effect of Different Tasks on Coherence Scoring}
This section examines how fine-tuning on diverse subsets of coherence proxy tasks affects coherence scoring. We fine-tune models on various combinations of these tasks, then perform final fine-tuning and evaluation on the coherence scoring task.

\begin{figure*}[t]
\centering
\includegraphics[width=0.98\textwidth]{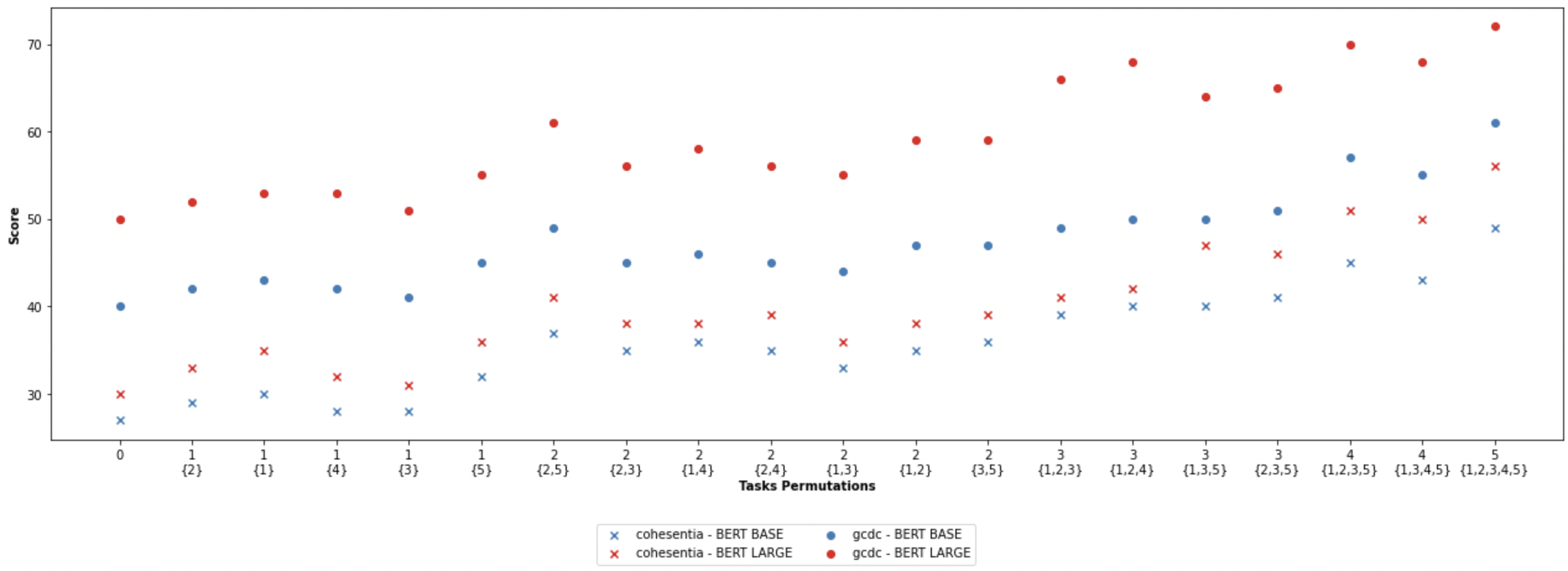}
\caption{Accuracy for Coherence Scoring Task for both GCDC and CoheSentia with different proxy coherence task-subsets. The labels are tasks IDs (1-SRO, 2-ISR, 3-DRR, 4-NPE, 5-NLI)}
\label{Fig:coherenceClassResults}
\vspace{-10pt}
\end{figure*}

Figure~\ref{Fig:coherenceClassResults} shows the impact of fine-tuning proxy coherence tasks on coherence scoring performance. Models fine-tuned on any coherence proxy task outperform those without fine-tuning (Ours-None), highlighting their effectiveness. Performance generally improves with more tasks, especially beyond three, indicating cumulative benefits.




Interestingly, NLI fine-tuning significantly enhances performance, likely due to its role in improving the model's ability to capture consistency, crucial for coherence assessment.
%
Additionally, ISR fine-tuning is more impactful when combined with other tasks. These findings underscore the importance of task selection and task interaction during fine-tuning for optimal coherence scoring.

\subsection{Impact of Non-Coherence Tasks Fine-Tuning}
\label{Par:CoherenceOnNoTCond}

In this Section we aim to empirically refute a possible hypothesis that the joint ALL model outperforms the NONE model simply due to its complexity, regardless of the nature of the tasks used (i.e., tasks reflecting coherence conditions). 

To this end, we compare the performance of our fine-tuned on coherence-tasks model (Ours-ALL) with a model fine-tuned on three tasks orthogonal to coherence, followed by fine-tuning on the coherence scoring task:
    (i) Machine Translation (MT): 
    We sample 15k instances from the WMT14 dataset \cite{bojar-EtAl:2014:W14-33}.
    (ii) Named-Entity Recognition (NER): 
    We use the Conll2003 dataset \cite{tjong-kim-sang-de-meulder-2003-introduction} containing annotations for 14k instances. 
    (iii) Part-of-Speech (POS), 
    using the same Conll2003 dataset containing the POS tags as well. 

\paragraph{Model}

In these experiments, we used the T5-large model as the basis, employing specific prompt and output designs for each task. For the NER and POS tasks, we adapted the ``Sentinel + Tag'' architecture by \citet{raman2022transforming}. Detailed prompts and sample outputs are provided in Appendix~\ref{App:T5Prompts}.

\paragraph{Results}
Following the same procedure as in our main experiments, fine-tuned models were assessed for coherence scoring using the GCDC and CoheSentia benchmarks (detailed in Table~\ref{Tab:CoherenceResults}). 


Fine-tuning on tasks orthogonal to coherence yielded minimal to no improvements over the baseline (Ours-None) and significantly underperformed compared to our final MTL model (Ours-ALL). This highlights the importance of coherence-specific proxy tasks for effective coherence detection, as unrelated tasks can hinder performance.

\subsection{Cross-Domain Generalization}
Since GCDC and CoheSentia present different domains and writing styles, we evaluate model generalizability by fine-tuning on one dataset and assessing coherence on the other.
Table~\ref{Tab:CoherenceResultsGeneralization} presents the results for our MTL model (Ours-ALL) and the non-coherence fine-tuned model (Ours-None) under three settings: fine-tuning on CoheSentia only, GCDC only, and both combined.




Results demonstrate performance gains across domains, highlighting the generalizability of our method. Combining data improves performance, with Ours-ALL showing a 12\% and 14\% error reduction on CoheSentia and GCDC, respectively, compared to in-domain scenarios, underscoring the utility and transferability of the learned features.

\begin{table}[t]
    \centering
    \scalebox{0.7}{
    \begin{tabular}{|m{10em}||c|c|}
    \hline
         Model                  & GCDC          & CoheSentia\\
         \hline
         \hline
         Ours-None-CoheSentia   & 52.8          & 34.8 \\
         Ours-None-GCDC         & 56.3          & 28.5 \\
         Ours-None-Both         & 57.5          & 35.4 \\
         \hline
         Ours-ALL-CoheSentia    & 71.8          & 62.3 \\
         Ours-ALL-GCDC          & 76.4          & 59.5 \\
         Ours-ALL-Both          & \textbf{79.8} & \textbf{66.7} \\
         \hline
    \end{tabular}}
    \caption{Accuracy on coherence scoring on both datasets when fine-tuned based on T5-model on only one dataset}
    \vspace{-10pt}
    \label{Tab:CoherenceResultsGeneralization}
\end{table}

\subsection{Effects of Different Tasks on One Another}

We investigate the impact of fine-tuning on various coherence task subsets on individual task performance. The model was trained on different task combinations with increasing numbers of tasks and evaluated on each task separately. 
%
%

\begin{figure}[t]
\centering
\includegraphics[width=0.5\textwidth]{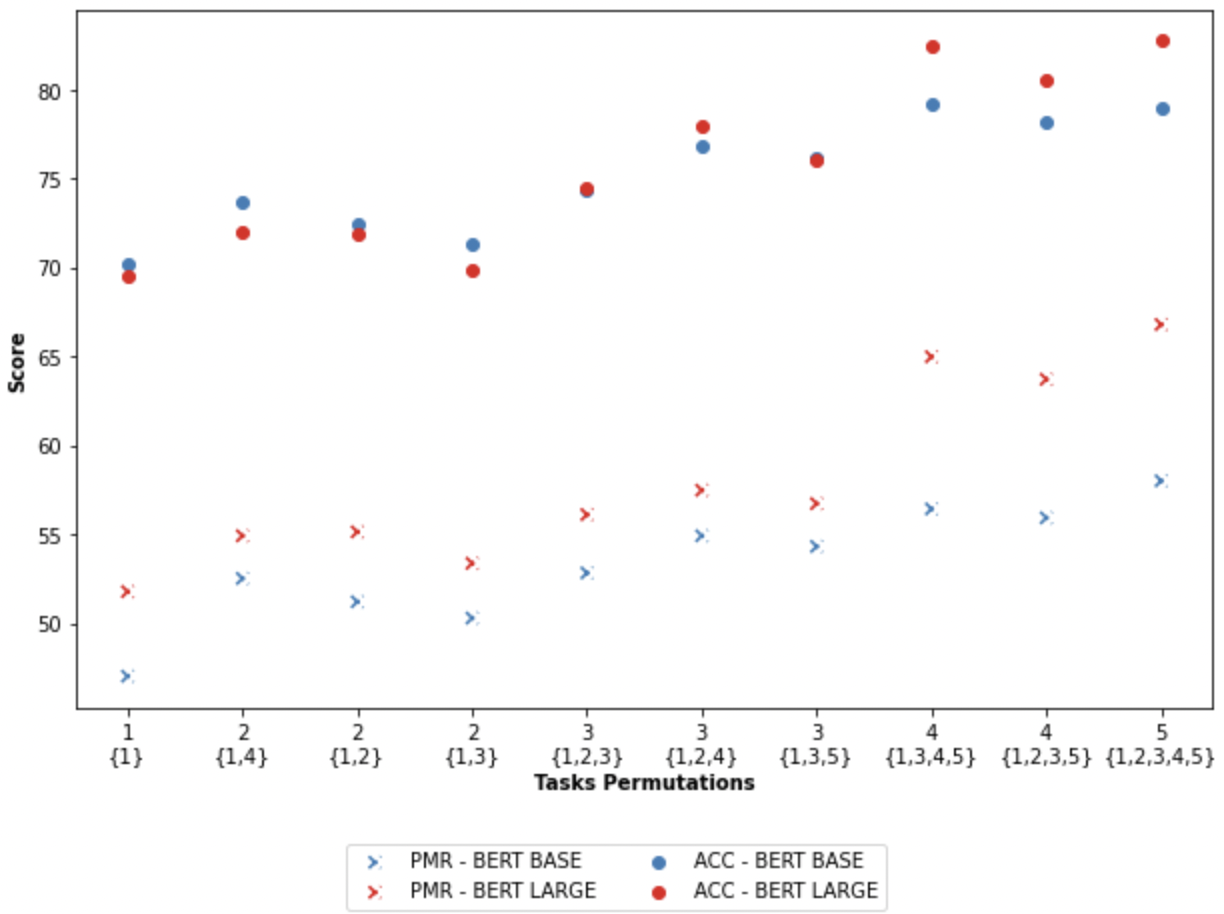}
\caption{Results for SRO task, for different subsets of coherence tasks fine-tuned upon. The labels are the number of tasks and in curly brackets which tasks (1 - SRO, 2 - ISR, 3 - DRR, 4 - NPE, 5 - NLI)}
\label{Fig:ReorderResOnly}
\vspace{-12pt}
\end{figure}

Figure~\ref{Fig:ReorderResOnly} shows consistent performance gains in the Sentence Reordering (SRO) task for BERT models as more tasks are jointly fine-tuned (see Appendix~\ref{App:PermutationFullRestult} for other tasks). This supports our hypothesis that shared information among coherence tasks enhances individual task performance.


The impact of specific tasks varies; for instance, DRR minimally affects SRO, likely due to limited training data. Notably, NLI significantly influences the performance of various tasks.
The ISR task notably improves performance on other tasks, suggesting its effectiveness in capturing relevance errors, crucial for coherence assessment. We thus emphasize the introduction of this self-supervised ISR task and advocate for its exploration in future research to enhance coherence assessment.

The overall performance trends are similar for both BERT-base and BERT-large models, indicating that the impact of specific tasks is consistent regardless of model size.

\section{Conclusion and Future Work}\label{Sec:Conclusion}

In this paper we propose a new coherence modeling method, based on \citet{Reinhart:1980}'s theory which defines the conditions needed for coherence: cohesion, consistency, and relevance. 
We use five key NLP tasks as proxies for these conditions, and train an MTL model on them jointly.
Our unified coherence model achieves SOTA results on these individual tasks, and moreover it excels in coherence scoring for both real-world and generated texts. 
We conjecture that this framework will enhance NLP systems' ability to quantify and 
evaluate text quality automatically. 
Future follow-up research will focus on using these conditions for improved coherent-text generation, and for detecting particular causes of incoherence automatically. Our code and models are publicly available, to encourage further research on {\em coherence scoring} and {\em coherence reasoning}.

\section*{Limitations}
While this work advances the modeling and automatic evaluation of coherence, limitations exist that suggest promising avenues for future research.
\paragraph{Dataset Limitations}
Existing coherence evaluation datasets like GCDC and CoheSentia, along with datasets for our proxy tasks, primarily focus on relatively short texts.
To address this, we analyzed the performance of our MTL models (Ours-ALL) and the non-coherence version (Ours-None) on GCDC and CoheSentia across various text lengths after fine-tuning for coherence scoring (see Figure~\ref{Fig:CoherenceAccuracyLength} in the Appendix). As expected, for both models and datasets, accuracy decreased with longer texts, highlighting the increased difficulty of assigning coherence scores for complex passages.
This observation aligns with recent work suggesting that while LLMs can handle longer texts, their reasoning abilities might decline with increasing text length \cite{levy2024task}.

\paragraph{Focus on Short Texts}
Our current study focused on short texts ($\leq512$ tokens). The effectiveness of our approach on longer documents remains an open question for future exploration.
We hypothesize that incorporating coherence proxy tasks could benefit the model's performance on longer texts, but further investigation is necessary.

\section*{Acknowledgements}

\bibliography{custom}

\begin{thebibliography}{59}
\providecommand{\natexlab}[1]{#1}

\bibitem[{Bai and Zhao(2018)}]{bai2018deep}
Hongxiao Bai and Hai Zhao. 2018.
\newblock \href {https://arxiv.org/abs/1807.05154} {Deep enhanced representation for implicit discourse relation recognition}.
\newblock \emph{Preprint}, arXiv:1807.05154.

\bibitem[{Barzilay and Lapata(2008)}]{Barzilay:2008}
Regina Barzilay and Mirella Lapata. 2008.
\newblock Modeling local coherence: An entity-based approach.
\newblock \emph{Computational Linguistics}, 34(1):1--34.

\bibitem[{Basu Roy~Chowdhury et~al.(2021)Basu Roy~Chowdhury, Brahman, and Chaturvedi}]{basu-roy-chowdhury-etal-2021-everything}
Somnath Basu Roy~Chowdhury, Faeze Brahman, and Snigdha Chaturvedi. 2021.
\newblock \href {https://doi.org/10.18653/v1/2021.emnlp-main.841} {Is everything in order? a simple way to order sentences}.
\newblock In \emph{Proceedings of the 2021 Conference on Empirical Methods in Natural Language Processing}, pages 10769--10779, Online and Punta Cana, Dominican Republic. Association for Computational Linguistics.

\bibitem[{Bojar et~al.(2014)Bojar, Buck, Federmann, Haddow, Koehn, Leveling, Monz, Pecina, Post, Saint-Amand, Soricut, Specia, and Tamchyna}]{bojar-EtAl:2014:W14-33}
Ondrej Bojar, Christian Buck, Christian Federmann, Barry Haddow, Philipp Koehn, Johannes Leveling, Christof Monz, Pavel Pecina, Matt Post, Herve Saint-Amand, Radu Soricut, Lucia Specia, and Ale~{s} Tamchyna. 2014.
\newblock \href {http://www.aclweb.org/anthology/W/W14/W14-3302} {Findings of the 2014 workshop on statistical machine translation}.
\newblock In \emph{Proceedings of the Ninth Workshop on Statistical Machine Translation}, pages 12--58, Baltimore, Maryland, USA. Association for Computational Linguistics.

\bibitem[{Bowman et~al.(2015)Bowman, Angeli, Potts, and Manning}]{bowman2015large}
Samuel~R. Bowman, Gabor Angeli, Christopher Potts, and Christopher~D. Manning. 2015.
\newblock \href {https://arxiv.org/abs/1508.05326} {A large annotated corpus for learning natural language inference}.
\newblock \emph{Preprint}, arXiv:1508.05326.

\bibitem[{Caruana(1997)}]{caruana1997multitask}
Rich Caruana. 1997.
\newblock Multitask learning.
\newblock \emph{Machine learning}, 28:41--75.

\bibitem[{Chen et~al.(2016)Chen, Qiu, and Huang}]{Chen:2016}
Xinchi Chen, Xipeng Qiu, and Xuanjing Huang. 2016.
\newblock \href {https://arxiv.org/abs/1607.06952} {Neural sentence ordering}.
\newblock \emph{Preprint}, arXiv:1607.06952.

\bibitem[{Dagan et~al.(2005)Dagan, Glickman, and Magnini}]{inproceedings}
Ido Dagan, Oren Glickman, and Bernardo Magnini. 2005.
\newblock \href {https://doi.org/10.1007/11736790_9} {The pascal recognising textual entailment challenge}.
\newblock pages 177--190.

\bibitem[{Devlin et~al.(2019)Devlin, Chang, Lee, and Toutanova}]{devlin-etal-2019-bert}
Jacob Devlin, Ming-Wei Chang, Kenton Lee, and Kristina Toutanova. 2019.
\newblock \href {https://doi.org/10.18653/v1/N19-1423} {{BERT}: Pre-training of deep bidirectional transformers for language understanding}.
\newblock In \emph{Proceedings of the 2019 Conference of the North {A}merican Chapter of the Association for Computational Linguistics: Human Language Technologies, Volume 1 (Long and Short Papers)}, pages 4171--4186, Minneapolis, Minnesota. Association for Computational Linguistics.

\bibitem[{Dijk(1979)}]{Dijk:1979}
Teun A~Van Dijk. 1979.
\newblock Pragmatic connectives.
\newblock \emph{Discourse processes}, 3:447--456.

\bibitem[{Dozat and Manning(2017)}]{dozat2017deep}
Timothy Dozat and Christopher~D. Manning. 2017.
\newblock \href {https://arxiv.org/abs/1611.01734} {Deep biaffine attention for neural dependency parsing}.
\newblock \emph{Preprint}, arXiv:1611.01734.

\bibitem[{Elazar et~al.(2022)Elazar, Basmov, Goldberg, and Tsarfaty}]{elazar2022textbased}
Yanai Elazar, Victoria Basmov, Yoav Goldberg, and Reut Tsarfaty. 2022.
\newblock \href {https://arxiv.org/abs/2109.12085} {Text-based np enrichment}.
\newblock \emph{Preprint}, arXiv:2109.12085.

\bibitem[{Feng et~al.(2014)Feng, Lin, , and Hirst}]{Feng:2014}
Vanessa~Wei Feng, Ziheng Lin, , and Graeme Hirst. 2014.
\newblock \href {https://www.aclweb.org/anthology/C14-1089} {The impact of deep hierarchical discourse structures in the evaluation of text coherence}.
\newblock \emph{Proceedings of COLING 2014, the 25th International Conference on Computational Linguistics: Technical Papers}, 1:940--949.

\bibitem[{Freeman and Gathercole(1966)}]{freeman1966perseveration}
Thomas Freeman and CE~Gathercole. 1966.
\newblock Perseveration—the clinical symptoms—in chronic schizophrenia and organic dementia.
\newblock \emph{The British Journal of Psychiatry}, 112(482):27--32.

\bibitem[{Givon(1995)}]{Givon:1995}
T~Givon. 1995.
\newblock Coherence in text vs. coherence in mind.
\newblock \emph{Coherence in spontaneous text}, pages 31--59.

\bibitem[{Goodfellow et~al.(2014)Goodfellow, Mirza, Da, Courville, and Bengio}]{DBLP:journals/corr/GoodfellowMDCB13}
Ian~J. Goodfellow, Mehdi Mirza, Xia Da, Aaron~C. Courville, and Yoshua Bengio. 2014.
\newblock \href {http://arxiv.org/abs/1312.6211} {An empirical investigation of catastrophic forgeting in gradient-based neural networks}.
\newblock In \emph{2nd International Conference on Learning Representations, {ICLR} 2014, Banff, AB, Canada, April 14-16, 2014, Conference Track Proceedings}.

\bibitem[{Guan et~al.(2021)Guan, Mao, Fan, Liu, Ding, and Huang}]{guan2021long}
Jian Guan, Xiaoxi Mao, Changjie Fan, Zitao Liu, Wenbiao Ding, and Minlie Huang. 2021.
\newblock \href {https://arxiv.org/abs/2105.08963} {Long text generation by modeling sentence-level and discourse-level coherence}.
\newblock \emph{Preprint}, arXiv:2105.08963.

\bibitem[{Guinaudeau and Strube(2013)}]{guinaudeau-strube-2013-graph}
Camille Guinaudeau and Michael Strube. 2013.
\newblock \href {https://aclanthology.org/P13-1010} {Graph-based local coherence modeling}.
\newblock In \emph{Proceedings of the 51st Annual Meeting of the Association for Computational Linguistics (Volume 1: Long Papers)}, pages 93--103, Sofia, Bulgaria. Association for Computational Linguistics.

\bibitem[{Halliday and Hasan(1976)}]{Halliday:1976}
M.A.K. Halliday and Ruqaiya Hasan. 1976.
\newblock \emph{Cohesion in English}.
\newblock Longman, Dallas, Texas.

\bibitem[{Hobbs(1979)}]{Hobbs:1979}
Jerry~R Hobbs. 1979.
\newblock Coherence and coreference.
\newblock \emph{Cognitive science}, 3(1):67--90.

\bibitem[{Honovich et~al.(2021)Honovich, Choshen, Aharoni, Neeman, Szpektor, and Abend}]{honovich2021q2}
Or~Honovich, Leshem Choshen, Roee Aharoni, Ella Neeman, Idan Szpektor, and Omri Abend. 2021.
\newblock \href {https://arxiv.org/abs/2104.08202} {$q^{2}$: Evaluating factual consistency in knowledge-grounded dialogues via question generation and question answering}.
\newblock \emph{Preprint}, arXiv:2104.08202.

\bibitem[{Hou(2021)}]{hou-2021-end-end}
Yufang Hou. 2021.
\newblock \href {https://doi.org/10.18653/v1/2021.findings-emnlp.119} {End-to-end neural information status classification}.
\newblock In \emph{Findings of the Association for Computational Linguistics: EMNLP 2021}, pages 1377--1388, Punta Cana, Dominican Republic. Association for Computational Linguistics.

\bibitem[{Ji and Eisenstein(2015)}]{ji-eisenstein-2015-one}
Yangfeng Ji and Jacob Eisenstein. 2015.
\newblock \href {https://doi.org/10.1162/tacl_a_00142} {One vector is not enough: Entity-augmented distributed semantics for discourse relations}.
\newblock \emph{Transactions of the Association for Computational Linguistics}, 3:329--344.

\bibitem[{Joshi and Weinstein(1981)}]{Joshi:1981}
Aravind~K Joshi and Scott Weinstein. 1981.
\newblock \href {https://aclanthology.org/J95-2003} {Control of inference: Role of some aspects of discourse structure-centering}.
\newblock \emph{IJCAI}, pages 385--387.

\bibitem[{Laban et~al.(2021)Laban, Dai, Bandarkar, and Hearst}]{laban2021transformer}
Philippe Laban, Luke Dai, Lucas Bandarkar, and Marti~A. Hearst. 2021.
\newblock \href {https://arxiv.org/abs/2107.03448} {Can transformer models measure coherence in text? re-thinking the shuffle test}.
\newblock \emph{Preprint}, arXiv:2107.03448.

\bibitem[{Lai and Tetreault(2018)}]{Lai:2018}
Alice Lai and Joel Tetreault. 2018.
\newblock \href {https://www.aclweb.org/anthology/W18-5023} {Discourse coherence in the wild: A dataset, evaluation and methods}.
\newblock \emph{In Proceedings of the 19th Annual SIGdial Meeting on Discourse and Dialogue. Association for Computational Linguistics}, 1:214--223.

\bibitem[{Lapata(2003)}]{Lapata:2003}
Mirella Lapata. 2003.
\newblock Probabilistic text structuring: Experiments with sentence ordering.
\newblock \emph{EMNLPIJCNLP}, pages 2273--2283.

\bibitem[{Levy et~al.(2024)Levy, Jacoby, and Goldberg}]{levy2024task}
Mosh Levy, Alon Jacoby, and Yoav Goldberg. 2024.
\newblock \href {https://arxiv.org/abs/2402.14848} {Same task, more tokens: the impact of input length on the reasoning performance of large language models}.
\newblock \emph{Preprint}, arXiv:2402.14848.

\bibitem[{Liang et~al.(2020)Liang, Zhao, and Webber}]{liang-etal-2020-extending}
Li~Liang, Zheng Zhao, and Bonnie Webber. 2020.
\newblock \href {https://doi.org/10.18653/v1/2020.codi-1.14} {Extending implicit discourse relation recognition to the {PDTB}-3}.
\newblock In \emph{Proceedings of the First Workshop on Computational Approaches to Discourse}, pages 135--147, Online. Association for Computational Linguistics.

\bibitem[{Lin et~al.(2011)Lin, Ng, , and Kan}]{Lin:2011}
Ziheng Lin, Hwee~Tou Ng, , and Min-Yen Kan. 2011.
\newblock \href {https://www.aclweb.org/anthology/P11-1100} {Automatically evaluating text coherence using discourse relations}.
\newblock \emph{Proceedings of the 49th Annual Meeting of the Association for Computational Linguistics: Human Language Technologies. Association for Computational Linguistics}, 1:997--1006.

\bibitem[{Liu et~al.(2023)Liu, Fu, and Strube}]{liu2023modeling}
Wei Liu, Xiyan Fu, and Michael Strube. 2023.
\newblock \href {https://arxiv.org/abs/2306.06472} {Modeling structural similarities between documents for coherence assessment with graph convolutional networks}.
\newblock \emph{Preprint}, arXiv:2306.06472.

\bibitem[{Liu et~al.(2020)Liu, Ou, Song, and Jiang}]{liu2020importance}
Xin Liu, Jiefu Ou, Yangqiu Song, and Xin Jiang. 2020.
\newblock \href {https://arxiv.org/abs/2004.12617} {On the importance of word and sentence representation learning in implicit discourse relation classification}.
\newblock \emph{Preprint}, arXiv:2004.12617.

\bibitem[{Logeswaran et~al.(2017)Logeswaran, Lee, and Radev}]{Logeswaran:2018}
Lajanugen Logeswaran, Honglak Lee, and Dragomir Radev. 2017.
\newblock \href {https://arxiv.org/abs/1611.02654} {Sentence ordering and coherence modeling using recurrent neural networks}.
\newblock \emph{Preprint}, arXiv:1611.02654.

\bibitem[{Long and Webber(2023)}]{long2023facilitating}
Wanqiu Long and Bonnie Webber. 2023.
\newblock \href {https://arxiv.org/abs/2301.02724} {Facilitating contrastive learning of discourse relational senses by exploiting the hierarchy of sense relations}.
\newblock \emph{Preprint}, arXiv:2301.02724.

\bibitem[{Maimon and Tsarfaty(2023)}]{anonymous:underreview}
Aviya Maimon and Reut Tsarfaty. 2023.
\newblock \href {https://doi.org/10.48550/ARXIV.2310.16329} {Cohesentia: {A} novel benchmark of incremental versus holistic assessment of coherence in generated texts}.
\newblock \emph{CoRR}, abs/2310.16329.

\bibitem[{Mann and Thompson(1988)}]{Mann:1988}
William~C Mann and Sandra~A Thompson. 1988.
\newblock Rhetorical structure theory: Toward a functional theory of text organization.
\newblock \emph{Text Interdisciplinary Journal for the Study of Discourse}, 1:243--281.

\bibitem[{Mesgar and Strube(2016)}]{mesgar-strube-2016-lexical}
Mohsen Mesgar and Michael Strube. 2016.
\newblock \href {https://doi.org/10.18653/v1/N16-1167} {Lexical coherence graph modeling using word embeddings}.
\newblock In \emph{Proceedings of the 2016 Conference of the North {A}merican Chapter of the Association for Computational Linguistics: Human Language Technologies}, pages 1414--1423, San Diego, California. Association for Computational Linguistics.

\bibitem[{Mesgar and Strube(2018)}]{mesgar-strube-2018-neural}
Mohsen Mesgar and Michael Strube. 2018.
\newblock \href {https://doi.org/10.18653/v1/D18-1464} {A neural local coherence model for text quality assessment}.
\newblock In \emph{Proceedings of the 2018 Conference on Empirical Methods in Natural Language Processing}, pages 4328--4339, Brussels, Belgium. Association for Computational Linguistics.

\bibitem[{Miltsakaki et~al.(2000)Miltsakaki, Eleni, Kukich, and Karen}]{Miltsakaki:2000}
Miltsakaki, Eleni, Kukich, and Karen. 2000.
\newblock Automated evaluation of coherence in student essays.

\bibitem[{Miltsakaki et~al.(2004)Miltsakaki, Prasad, Joshi, and Webber}]{miltsakaki-etal-2004-penn}
Eleni Miltsakaki, Rashmi Prasad, Aravind Joshi, and Bonnie Webber. 2004.
\newblock \href {http://www.lrec-conf.org/proceedings/lrec2004/pdf/618.pdf} {The {P}enn {D}iscourse {T}reebank}.
\newblock In \emph{Proceedings of the Fourth International Conference on Language Resources and Evaluation ({LREC}{'}04)}, Lisbon, Portugal. European Language Resources Association (ELRA).

\bibitem[{Mostafazadeh et~al.(2016)Mostafazadeh, Chambers, He, Parikh, Batra, Vanderwende, Kohli, and Allen}]{mostafazadeh2016corpus}
Nasrin Mostafazadeh, Nathanael Chambers, Xiaodong He, Devi Parikh, Dhruv Batra, Lucy Vanderwende, Pushmeet Kohli, and James Allen. 2016.
\newblock \href {https://arxiv.org/abs/1604.01696} {A corpus and evaluation framework for deeper understanding of commonsense stories}.
\newblock \emph{Preprint}, arXiv:1604.01696.

\bibitem[{Pitler et~al.(2009)Pitler, Louis, and Nenkova}]{pitler-etal-2009-automatic}
Emily Pitler, Annie Louis, and Ani Nenkova. 2009.
\newblock \href {https://aclanthology.org/P09-1077} {Automatic sense prediction for implicit discourse relations in text}.
\newblock In \emph{Proceedings of the Joint Conference of the 47th Annual Meeting of the {ACL} and the 4th International Joint Conference on Natural Language Processing of the {AFNLP}}, pages 683--691, Suntec, Singapore. Association for Computational Linguistics.

\bibitem[{Prasad et~al.(2008)Prasad, Dinesh, Lee, Miltsakaki, Robaldo, Joshi, and Webber}]{prasad-etal-2008-penn}
Rashmi Prasad, Nikhil Dinesh, Alan Lee, Eleni Miltsakaki, Livio Robaldo, Aravind Joshi, and Bonnie Webber. 2008.
\newblock \href {http://www.lrec-conf.org/proceedings/lrec2008/pdf/754_paper.pdf} {The {P}enn {D}iscourse {T}ree{B}ank 2.0.}
\newblock In \emph{Proceedings of the Sixth International Conference on Language Resources and Evaluation ({LREC}'08)}, Marrakech, Morocco. European Language Resources Association (ELRA).

\bibitem[{Prasad et~al.(2019)Prasad, Webber, Lee, and Joshi}]{AB2/SUU9CB_2019}
Rashmi Prasad, Bonnie Webber, Alan Lee, and Aravind Joshi. 2019.
\newblock \href {https://doi.org/11272.1/AB2/SUU9CB} {{Penn Discourse Treebank Version 3.0}}.

\bibitem[{Qin et~al.(2016)Qin, Zhang, and Zhao}]{qin-etal-2016-shallow}
Lianhui Qin, Zhisong Zhang, and Hai Zhao. 2016.
\newblock \href {https://doi.org/10.18653/v1/K16-2010} {Shallow discourse parsing using convolutional neural network}.
\newblock In \emph{Proceedings of the {C}o{NLL}-16 shared task}, pages 70--77, Berlin, Germany. Association for Computational Linguistics.

\bibitem[{Raffel et~al.(2020)Raffel, Shazeer, Roberts, Lee, Narang, Matena, Zhou, Li, and Liu}]{raffel2020exploring}
Colin Raffel, Noam Shazeer, Adam Roberts, Katherine Lee, Sharan Narang, Michael Matena, Yanqi Zhou, Wei Li, and Peter~J. Liu. 2020.
\newblock \href {https://arxiv.org/abs/1910.10683} {Exploring the limits of transfer learning with a unified text-to-text transformer}.
\newblock \emph{Preprint}, arXiv:1910.10683.

\bibitem[{Raman et~al.(2022)Raman, Naim, Chen, Hashimoto, Yalasangi, and Srinivasan}]{raman2022transforming}
Karthik Raman, Iftekhar Naim, Jiecao Chen, Kazuma Hashimoto, Kiran Yalasangi, and Krishna Srinivasan. 2022.
\newblock \href {https://arxiv.org/abs/2203.08378} {Transforming sequence tagging into a seq2seq task}.
\newblock \emph{Preprint}, arXiv:2203.08378.

\bibitem[{Reinhart(1980)}]{Reinhart:1980}
Tanya Reinhart. 1980.
\newblock \emph{Conditions for text coherence. Poetics Today 1(4): 16t-180}, volume~1.

\bibitem[{Shrimai~Prabhumoye(2020)}]{Prabhumoye:2020}
Alan W~Black Shrimai~Prabhumoye, Ruslan~Salakhutdinov. 2020.
\newblock \href {https://arxiv.org/pdf/2005.00432.pdf} {Topological sort for sentence ordering}.

\bibitem[{Somasundaran et~al.(2014)Somasundaran, Burstein, and Chodorow}]{Somasundaran:2014}
Swapna Somasundaran, Jill Burstein, and Martin Chodorow. 2014.
\newblock \href {https://www.aclweb.org/anthology/C14-1090} {Lexical chaining for measuring discourse coherence quality in test-taker essays}.
\newblock \emph{Proceedings of COLING 2014, the 25th International Conference on Computational Linguistics: Technical Papers}, 1:950--961.

\bibitem[{Tarjan(1976)}]{Tarjan:1976}
Robert~Endre Tarjan. 1976.
\newblock Edge-disjoint spanning trees and depth-first search.
\newblock \emph{Acta Informatica}, 6(2):171--185.

\bibitem[{Tay et~al.(2017)Tay, Phan, Tuan, and Hui}]{tay2017skipflow}
Yi~Tay, Minh~C. Phan, Luu~Anh Tuan, and Siu~Cheung Hui. 2017.
\newblock \href {https://arxiv.org/abs/1711.04981} {Skipflow: Incorporating neural coherence features for end-to-end automatic text scoring}.
\newblock \emph{Preprint}, arXiv:1711.04981.

\bibitem[{Tjong Kim~Sang and De~Meulder(2003)}]{tjong-kim-sang-de-meulder-2003-introduction}
Erik~F. Tjong Kim~Sang and Fien De~Meulder. 2003.
\newblock \href {https://aclanthology.org/W03-0419} {Introduction to the {C}o{NLL}-2003 shared task: Language-independent named entity recognition}.
\newblock In \emph{Proceedings of the Seventh Conference on Natural Language Learning at {HLT}-{NAACL} 2003}, pages 142--147.

\bibitem[{Wei et~al.(2023)Wei, Wang, Schuurmans, Bosma, Ichter, Xia, Chi, Le, and Zhou}]{wei2023chainofthought}
Jason Wei, Xuezhi Wang, Dale Schuurmans, Maarten Bosma, Brian Ichter, Fei Xia, Ed~Chi, Quoc Le, and Denny Zhou. 2023.
\newblock \href {https://arxiv.org/abs/2201.11903} {Chain-of-thought prompting elicits reasoning in large language models}.
\newblock \emph{Preprint}, arXiv:2201.11903.

\bibitem[{Williams et~al.(2018)Williams, Nangia, and Bowman}]{N18-1101}
Adina Williams, Nikita Nangia, and Samuel Bowman. 2018.
\newblock \href {http://aclweb.org/anthology/N18-1101} {A broad-coverage challenge corpus for sentence understanding through inference}.
\newblock In \emph{Proceedings of the 2018 Conference of the North American Chapter of the Association for Computational Linguistics: Human Language Technologies, Volume 1 (Long Papers)}, pages 1112--1122. Association for Computational Linguistics.

\bibitem[{Wolf et~al.(2020)Wolf, Debut, Sanh, Chaumond, Delangue, Moi, Cistac, Rault, Louf, Funtowicz, Davison, Shleifer, von Platen, Ma, Jernite, Plu, Xu, Scao, Gugger, Drame, Lhoest, and Rush}]{wolf2020huggingfaces}
Thomas Wolf, Lysandre Debut, Victor Sanh, Julien Chaumond, Clement Delangue, Anthony Moi, Pierric Cistac, Tim Rault, Rémi Louf, Morgan Funtowicz, Joe Davison, Sam Shleifer, Patrick von Platen, Clara Ma, Yacine Jernite, Julien Plu, Canwen Xu, Teven~Le Scao, Sylvain Gugger, Mariama Drame, Quentin Lhoest, and Alexander~M. Rush. 2020.
\newblock \href {https://arxiv.org/abs/1910.03771} {Huggingface's transformers: State-of-the-art natural language processing}.
\newblock \emph{Preprint}, arXiv:1910.03771.

\bibitem[{Xiang et~al.(2022)Xiang, Wang, Dai, and Mo}]{xiang-etal-2022-encoding}
Wei Xiang, Bang Wang, Lu~Dai, and Yijun Mo. 2022.
\newblock \href {https://doi.org/10.18653/v1/2022.findings-acl.256} {Encoding and fusing semantic connection and linguistic evidence for implicit discourse relation recognition}.
\newblock In \emph{Findings of the Association for Computational Linguistics: ACL 2022}, pages 3247--3257, Dublin, Ireland. Association for Computational Linguistics.

\bibitem[{Xu et~al.(2018)Xu, Ren, Zhang, Zeng, Cai, and Sun}]{xu2018skeletonbased}
Jingjing Xu, Xuancheng Ren, Yi~Zhang, Qi~Zeng, Xiaoyan Cai, and Xu~Sun. 2018.
\newblock \href {https://arxiv.org/abs/1808.06945} {A skeleton-based model for promoting coherence among sentences in narrative story generation}.
\newblock \emph{Preprint}, arXiv:1808.06945.

\bibitem[{Yi et~al.(2019)Yi, Goel, Khatri, Cervone, Chung, Hedayatnia, Venkatesh, Gabriel, and Hakkani-Tur}]{yi2019coherent}
Sanghyun Yi, Rahul Goel, Chandra Khatri, Alessandra Cervone, Tagyoung Chung, Behnam Hedayatnia, Anu Venkatesh, Raefer Gabriel, and Dilek Hakkani-Tur. 2019.
\newblock \href {https://arxiv.org/abs/1904.13015} {Towards coherent and engaging spoken dialog response generation using automatic conversation evaluators}.
\newblock \emph{Preprint}, arXiv:1904.13015.

\end{thebibliography}

\appendix

\section{Tasks Specific Experimental Settings} \label{App:TasksExperimentalSettings}
In this section, we further elaborate on the datasets and evaluation metrics used for each one of the coherence proxy tasks. 

\subsection{The Sentence Reordering Task Setup}
\paragraph{Topological Sort:}\label{App:topologicalSort}
A topological sort \cite{Tarjan:1976} linearly orders vertices in a DAG. The algorithm is presented in Algo~\ref{algo:topologicalSort}.


\begin{algorithm}
\caption{Topological Sort Algorithm} 
Input: A digraph G with n vertices

Output: A topological ordering v1,v2...vn of G.
\begin{algorithmic}
    \State L $\leftarrow$ Empty list that will contain the sorted nodes
    \State S $\leftarrow$ Set of all nodes with no incoming edge
    \While {S is not empty}
        \State remove a node n from S
        \State add n to L
        \For{each node m with an edge e from n to m}
            \State remove edge e from the graph
            \If {m has no other incoming edges}
                \State insert m into S
            \EndIf
        \EndFor
    \EndWhile
    \If {graph has edges}
        \State \Return error (graph has at least one cycle)
    \Else
        \State \Return L (a topologically sorted order)
    \EndIf 
\end{algorithmic}
\label{algo:topologicalSort}
\end{algorithm}


\paragraph{Dataset:} We use the ROCStories \cite{mostafazadeh2016corpus} dataset (Licence ID is CC-BY 4.0.) which contains 5-sentence stories. 
We use the standard 85:15 train/test split and randomly select a subset of the train for validation.

\paragraph{Evaluation:} We use two common evaluation metrics for the reordering  task:\footnote{There are 5 metrics, we used the most common 2.}
\begin{itemize}
    \item Perfect Match Ratio (PMR): \citet{Chen:2016} calculate the percentage of samples for which the entire sequence was correctly predicted.
\[PMR = \frac{1}{N}\sum_{i=1}^{N} 1\{\hat{O}^i = O^i\}\]
    \item Sentence Accuracy (Acc): \citet{Logeswaran:2018} calculate the percentage of sentences for which their absolute position was correctly predicted. 
    \[Acc = \frac{1}{N}\sum_{i=1}^{N}\frac{1}{v_i}\sum_{j=1}^{v_i} 1\{\hat{O}_j^i = O_j^{i}\}\]
\end{itemize}


\subsection{The Discourse-Relation Recognition Task Setup}
\paragraph{Dataset:} We use the Penn Discourse TreeBank3 (PDTB3) Level 2 dataset \cite{miltsakaki-etal-2004-penn, prasad-etal-2008-penn, liang-etal-2020-extending}. 
\rtRemove{
The PDTB is a resource for extracting discourse relations between pairs of sentences.
} 
We only used labels with more than 100 instances, which leaves us with 14 senses from $L_2$. 
The variability of data splits used in the literature is substantial, therefore, we follow earlier work by \citet{ji-eisenstein-2015-one, bai2018deep, liu2020importance, xiang-etal-2022-encoding} using Sections 2-20, 0-1 and 21-22 for training, validation and testing respectively.
When multiple annotated labels are present, we adopt the approach described by \citet{qin-etal-2016-shallow} and consider them as distinct instances during the training phase. During testing, if a prediction matches any of the reference labels, it is considered correct.
\rtRemove{
More on the PDTB dataset and possible discourse relations are in App~\ref{APP:PDTBDataset}.
}

\paragraph{Evaluation:} We use the accuracy metric on the number of sentence pairs the model correctly predicted the $L_2$ discourse relation:
\[Accuracy = \frac{1}{N}\sum_{i=1}^{N} 1\{\hat{R}^i = R^i\}\]

\subsection{The NP Enrichment Task Setup}
\begin{figure}[H]
\centering
\includegraphics[width=0.5\textwidth]{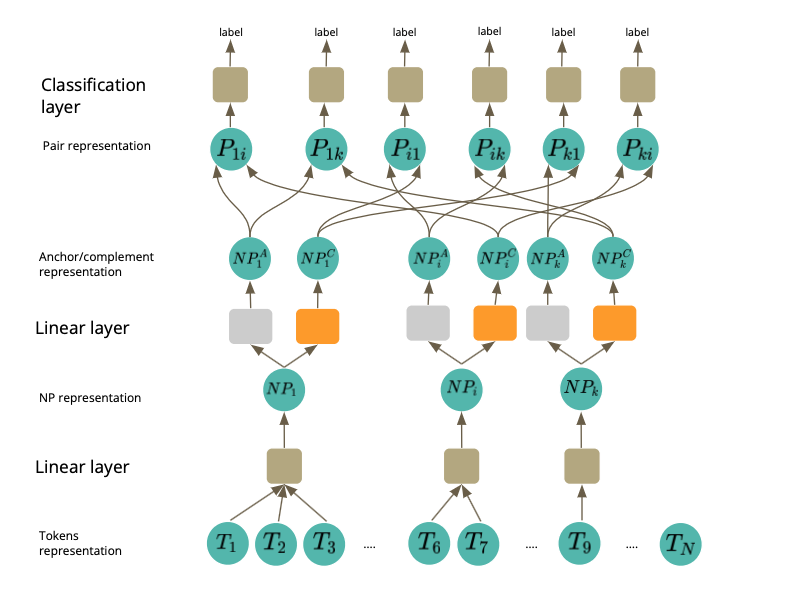}
\caption{Illustration of the token head which contains several stages: starting with (1) embedding for each token in the text, (2) creating an embedding for each NP when it acts as the complement and the anchor separately, (3) a representation for each NP pair and finally (4) a classification layer}
\label{Fig:tokenHead}
\end{figure}

\paragraph{Token Classification Head:}
Figure~\ref{Fig:tokenHead} is an illustration of the token classification head for the NPE task. 

\paragraph{Dataset:} We use the TNE dataset \cite{elazar2022textbased} (Licence Free) which contains documents and relations between every noun pair in it (with a total number of nouns of 190k and a total number of NP relations of 1M). There are 28 possible relations (including `no relation'). This dataset's advantage is that it contains real-world long paragraphs. 
As in the original publication split the data at the document level.
%

\begin{figure}[H]
    \centering
    \includegraphics[width=0.4\textwidth]{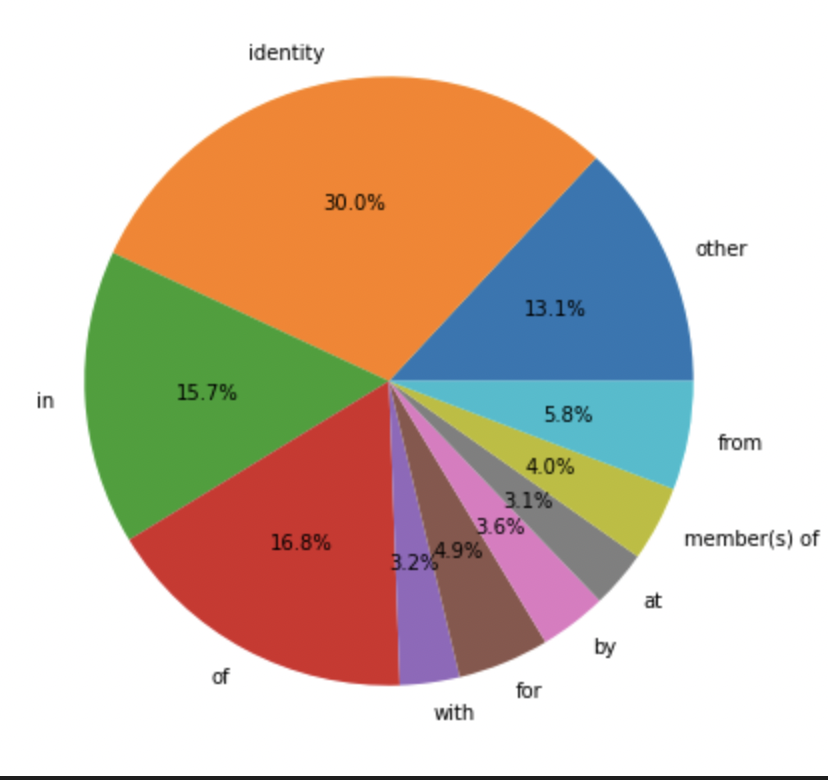}
    \caption{Distribution of the main prepositions in the NP Enrichment test set}
    \label{Fig:TNEDistribution}
\end{figure}

\rtRemove{
There are 28 unevenly distributed possible preposition relations between pair of nouns, among them: 'no-relation', 'identity-time/date/measurement', 'at', 'near', 'about', 'with', 'outside', 'during', 'for', 'between', 'member(s) of', 'in', 'of', 'against', 'under', 'identity-standard', 'by', 'from', 'before', 'around', 'over', 'into', 'to', 'among', 'after', 'inside', 'on', 'identity-idiomatic']. 
The distribution of them is in Figure~\ref{Fig:TNEDistribution}
}
The distribution of the possible preposition between pair of nouns in TNE dataset is in Figure~\ref{Fig:TNEDistribution}

\rtRemove{
In Figure~\ref{Fig:TNEExample} you can see an example of a text with all noun entities marked.
\begin{figure*}[h!]
    \centering
    \includegraphics[width=0.5\textwidth]{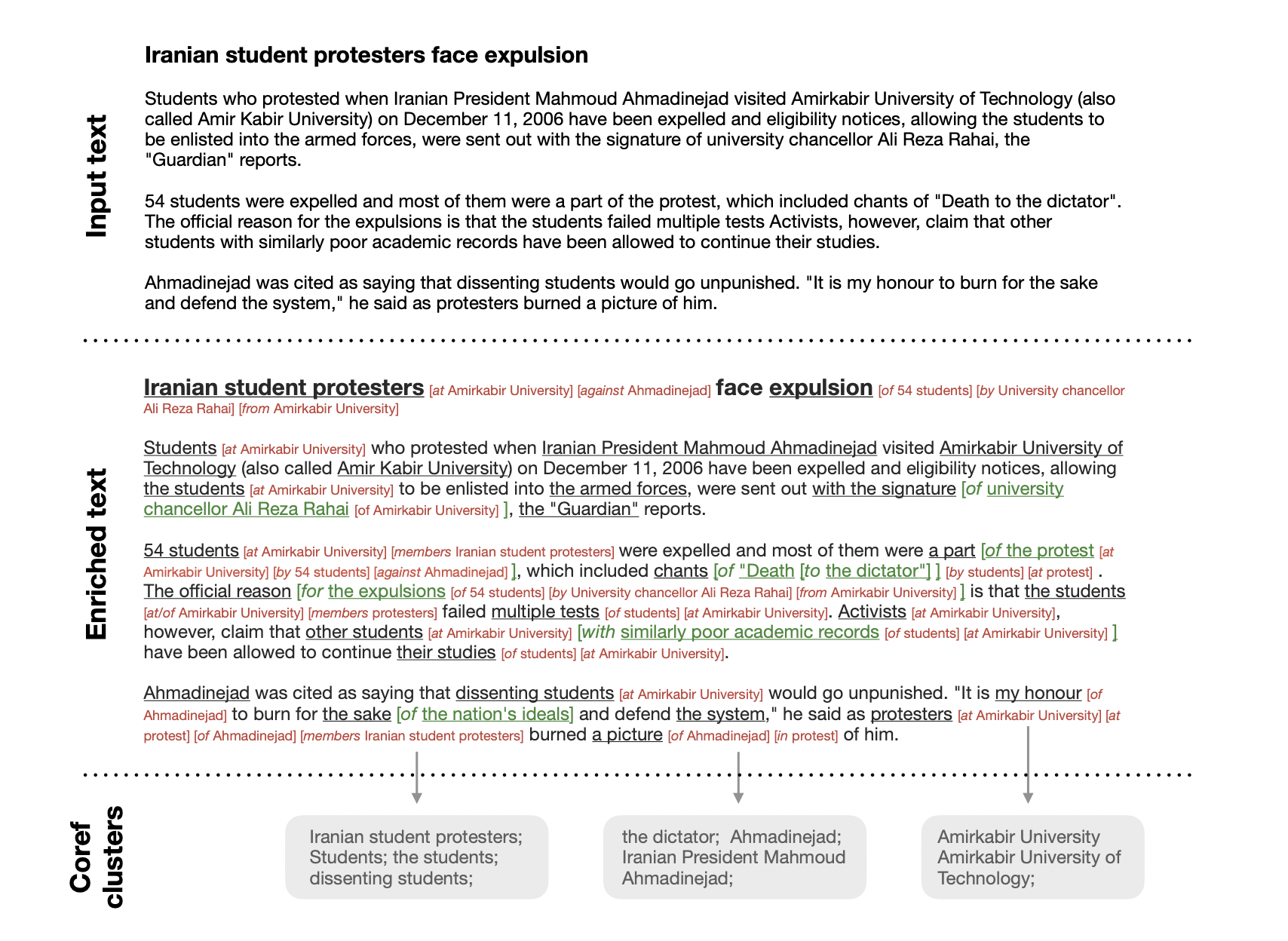}
    \caption{An example of an NP-enriched document from the TNE dataset. In the upper image there is the original paragraph, in the middle - is the document after enrichment (in green are the NP-enrichments explicitly mentioned, and in red are those that are not explicitly mentioned), at the bottom are the clusters for some of the NP's}
    \label{Fig:TNEExample}
\end{figure*}
}

\paragraph{Evaluation:} We report precision, recall \&  F1 on NP pairs with prepositional links between them.

\subsection{The NL Inference (NLI) Task Setup}
\paragraph{Dataset and Evaluation:}
We use the MNLI dataset \cite{N18-1101} (Licence ID CC-BY-3.0). 
 with the accuracy metric on the amount of hypothesis-premise pairs that the model correctly predicts their relation \(R\):
\[Accuracy = \frac{1}{N}\sum_{i=1}^{N} 1\{\hat{R}^i = R^i\}\]

\subsection{The Irrelevant Sentence Recognition  Task Setup}

\paragraph{Dataset:} We again use ROCStories as in sentence reordering. 
Each story within the ROCStories dataset was augmented with a single, randomly inserted sentence. 
The irrelevant sentence for each story was randomly selected from the entire ROCStories dataset, with the sole constraint that it contained entities present in the target story. 
Both this and the sentence reordering task leverage the same benchmark, retaining the same train/dev/test splits.

\paragraph{Evaluation:} We use the accuracy metric on the percentage of paragraphs where the model correctly detected the irrelevant sentence \(S\):
\[Accuracy = \frac{1}{N}\sum_{i=1}^{N} 1\{\hat{S}^i = S^i\}\]

\subsection{Overall Experimental Settings}
We trained each model three times, reporting the mean performance. Training utilized multiple Tesla V100 GPUs (up to 4) with 32GB memory each.
For each architecture, the settings are: 
\begin{enumerate}
    \item Classification-Based: BERT (base and large) served as the encoder with fine-tuning across all layers. We used Adam optimizer with a learning rate of 5e-5 and a dropout of 0.5. For tasks requiring classification (SRO, ISR, DRR, NLI), we employed a linear classification head with 512 hidden dimensions and 0.3 dropouts. The NPE  utilized a different head structure (details omitted for brevity). Cross-Entropy loss was used for all datasets.
    \item Generation-Based: T5 (base and large) models were used as the backbone. Training employed Adam optimizer with a learning rate of 5e-5. Models were trained with task-specific prompts and corresponding ground truth labels for supervised learning. 
\end{enumerate}
Both architectures shared the following hyper-parameters: fine-tuning for 3 epochs with early stopping, batch size of 4, and gradient accumulation steps of 2. The hyper-parameters were chosen using parameters-grid. Our code is based on the Huggingface library \cite{wolf2020huggingfaces}. 

\section{Coherence Assessment Experimental Settings}\label{App:CoherenceExperimentalSettings}
For each architecture, the settings are:
\begin{enumerate}
    \item Classification-Based (BERT base and large): Encoder with fine-tuning across all layers, Adam optimizer (learning rate 5e-4), dropout (0.3). Each dataset used a linear classification head (512 hidden dimensions, 0.1 dropout). Cross-Entropy loss was used.
    \item Generation-Based (T5 base and large): Encoder-decoder architecture, Adam optimizer (learning rate 2e-5). Inputs included prompts specific to each dataset (GCDC or CoheSentia) and the paragraph text.
\end{enumerate}

The models share hyperparameters: 50 epochs with early stopping (accuracy), batch size of 4, and gradient accumulation steps of 2. We employed 10-fold cross-validation on both datasets (following \citet{Lai:2018}) using a single Tesla V100 GPU with 32GB memory. The hyper-parameters were chosen using parameters-grid. Our code is based on the Huggingface library \cite{wolf2020huggingfaces}.

\section{Text Length vs. Coherence Score} \label{App:TextLength}
\begin{figure}[t]
\centering
    \includegraphics[width=0.5\textwidth]{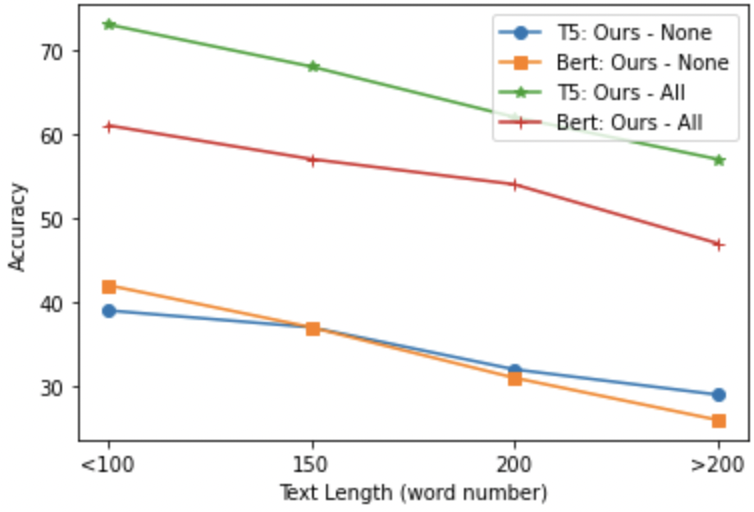}
    \caption{Accuracy For CoheSentia based on number of words}
    \label{Fig:cohesentialengthRes}
\hfill
\label{Fig:CoherenceAccuracyLength}
\end{figure}

The accuracy of the models on both coherence datasets based on different lengths is in Figure~\ref{Fig:CoherenceAccuracyLength}.

\begin{figure}[t]
\centering
    \includegraphics[width=0.5\textwidth]{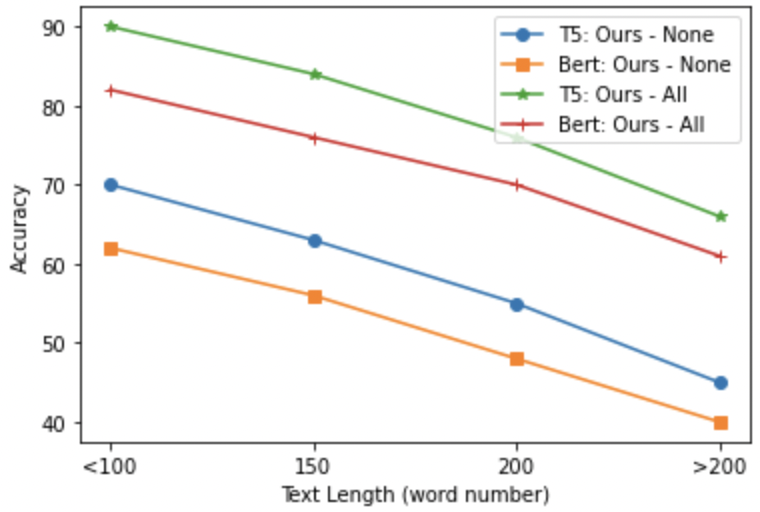}
    \caption{Accuracy For GCDC based on number of words}
    \label{Fig:gcdclengthRes}
\end{figure}

\section{Qualitative Analysis}\label{App:QualitativeExample}
\subsection{Qualitative Analysis}
To gain qualitative insights, we sampled 50 misclassified examples by SOTA models, 
from CoheSentia and GCDC. We then assessed these examples on various models, including our MTL model (Ours-ALL) and the non-coherence fine-tuning version (Ours-None).

\begin{figure*}[t]
\centering
\begin{subfigure}{0.8\textwidth}
    \includegraphics[width=\textwidth]{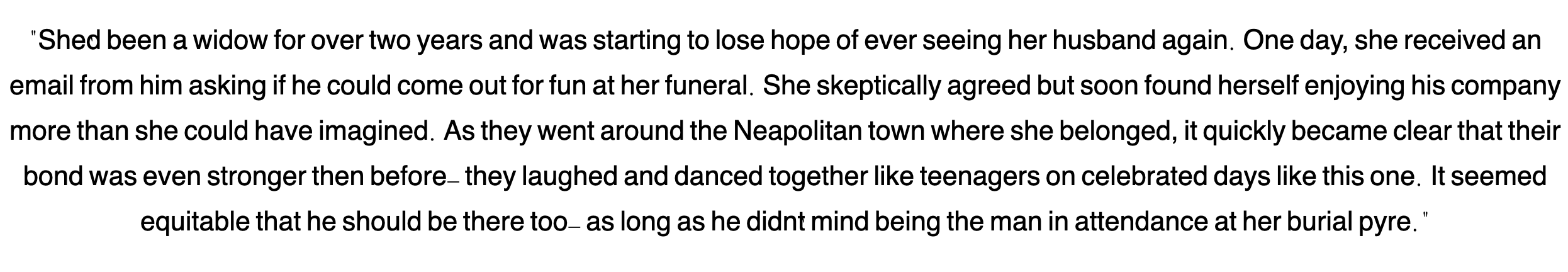}
    \caption{CoheSentia}
    \label{Fig:ExampleCoheSentia}
\end{subfigure}
\hfill
\begin{subfigure}{0.8\textwidth}
    \includegraphics[width=\textwidth]{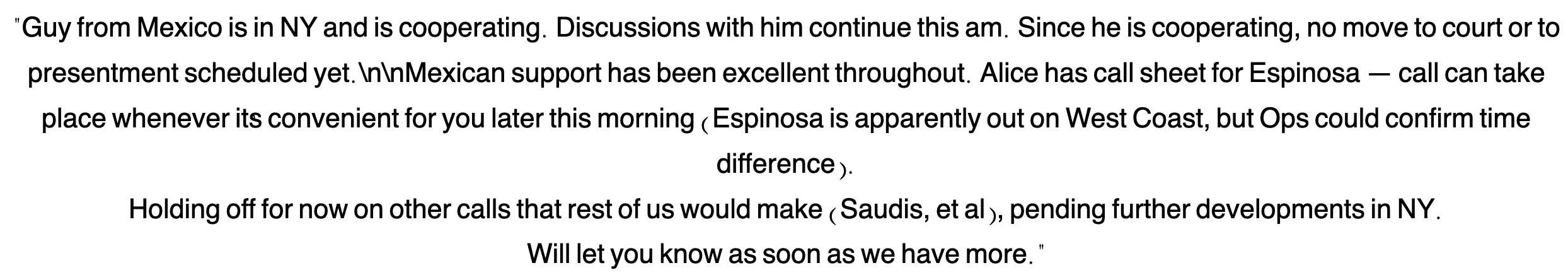}
    \caption{GCDC}
    \label{Fig:ExampleGCDC}
\end{subfigure}
\hfill
\caption{Sample Texts for coherence scoring tasks: GCDC \& CoheSentia benchmarks}
\label{Fig:CoherenceExamples}
\end{figure*}

For CoheSentia, the previous SOTA models favor extreme scores, likely due to training data imbalance. Our model exhibits greater robustness, predicting a more balanced distribution of scores.
Figure~\ref{Fig:ExampleCoheSentia} and Table~\ref{Tab:ExampleCoheSentia} present an example of a text from the CoheSentia dataset and the predictions of the models. 
In this example, the base model (Ours-None) failed on coherence prediction, while our final model (Ours-ALL) succeeded. 
Figure~\ref{Fig:ExampleGCDC} presents an example of text from GCDC dataset and Table~\ref{Tab:ExampleGCDC} the predictions of different models on the coherence scoring task. 
This example highlights a complex case with cohesion and relevance violations. Both the baseline and ISR-trained models missed this issue, while our MTL model achieved accurate prediction.

\begin{table}[t]
    \centering
    \scalebox{0.7}{
    \begin{tabular}{|c|c|}
        \hline
        \hline
        Model &  Prediction \\
        \hline
        Ground Truth            &  Medium \\
        \hline
        SOTA                    &  High \\
        Ours-None (BERT-large)  &  High \\
        Ours-None (T5-large)    &  High \\
        Ours-ALL (BERT-large)   &  Medium \\
        Ours-ALL (T5-large)     &  Medium \\
        \hline
        \hline
    \end{tabular}}
    \caption{Predicted Coherence scores for the text in Figure~\ref{Fig:ExampleCoheSentia}}
    \label{Tab:ExampleCoheSentia}
\end{table}

\begin{table}[t]
    \centering
    \scalebox{0.7}{
    \begin{tabular}{|c|c|}
        \hline
        \hline
        Model &  Prediction \\
        \hline
        Ground Truth            &  Low \\
        \hline
        SOTA                    &  Medium \\
        Ours-None (BERT-large)  &  Medium \\
        Ours-None (T5-large)    &  Medium \\
        Ours-ALL (BERT-large)   &  Low \\
        Ours-ALL (T5-large)     &  Low \\
        \hline
        \hline
    \end{tabular}}
    \caption{Predicted Coherence scores for the text in Figure~\ref{Fig:ExampleGCDC}}
    \label{Tab:ExampleGCDC}
\end{table}

\section{Results for Subsets of Tasks} \label{App:PermutationFullRestult}
\begin{figure}[t]
\centering
\begin{subfigure}{0.5\textwidth}
    \includegraphics[width=\textwidth]{images/reorder_results_class2.png}
    \caption{SRO}
    \label{Fig:ReorderRes}
\end{subfigure}
\hfill
\begin{subfigure}{0.5\textwidth}
    \includegraphics[width=\textwidth]{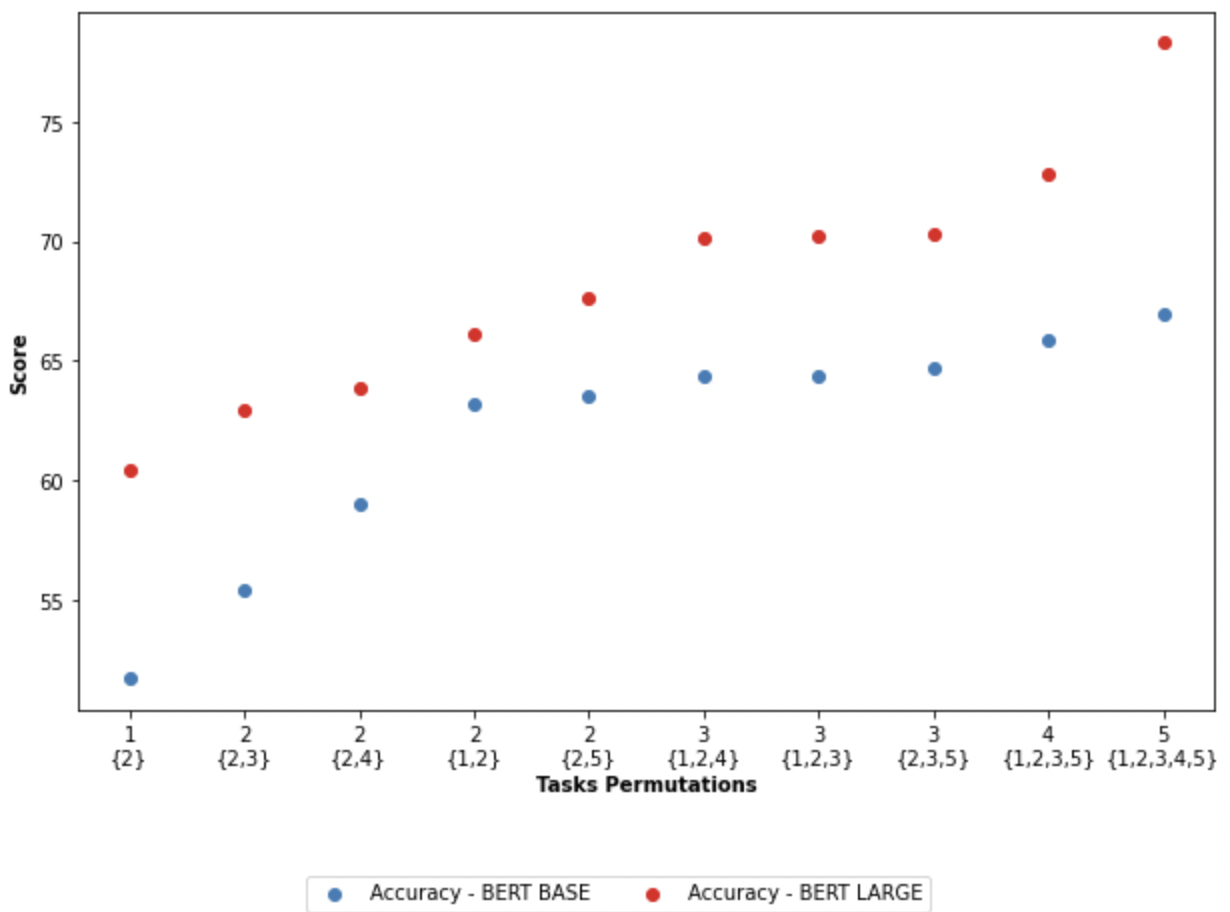}
    \caption{ISR}
    \label{Fig:RelevanceRes}
\end{subfigure}
\hfill
\begin{subfigure}{0.5\textwidth}
    \includegraphics[width=\textwidth]{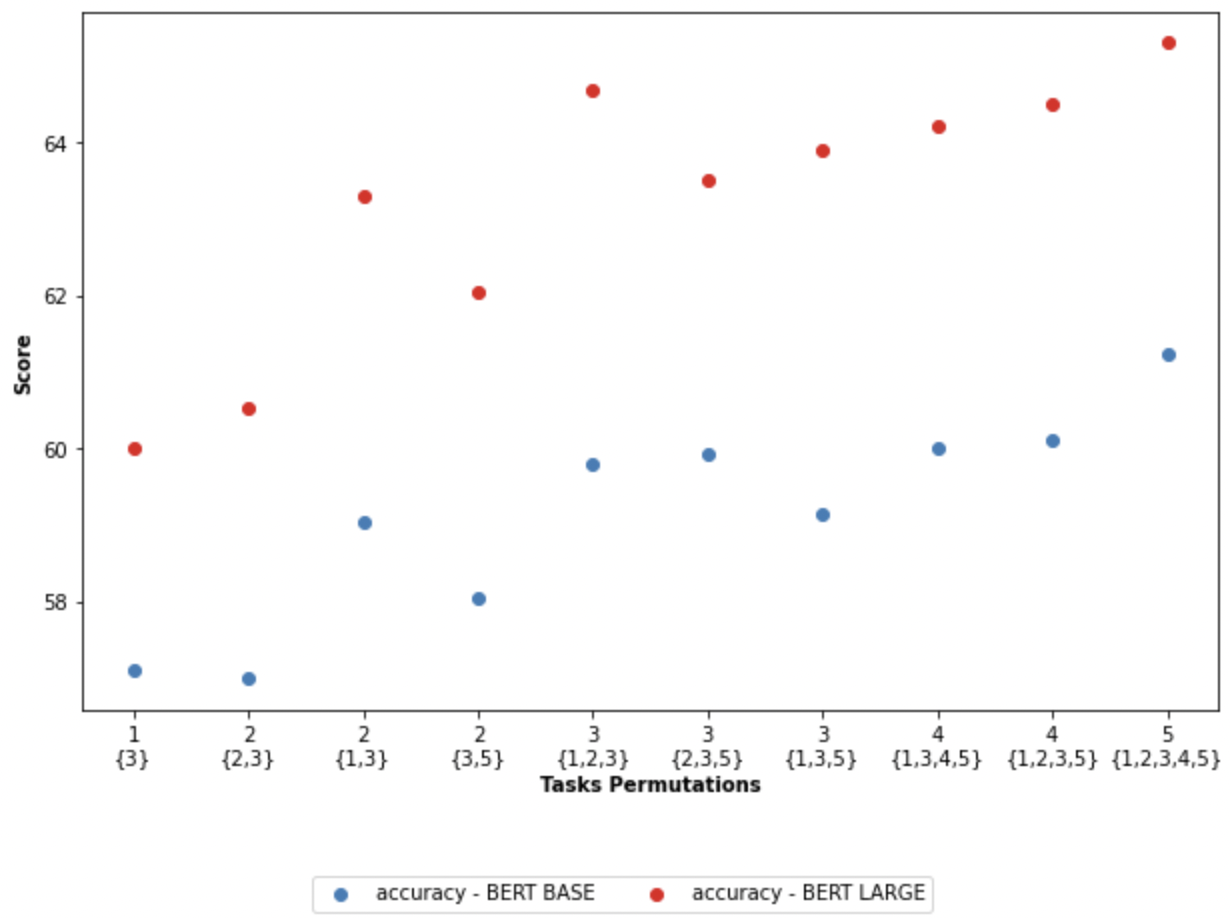}
    \caption{DRR}
    \label{Fig:DiscourseRes}
\end{subfigure}
\hfill
\caption{Results for all tasks, for different permutations of tasks fine-tuned upon. The labels are the number of tasks and in curly brackets which tasks (1 - SRO, 2 - ISR, 3 - DRR, 4 - NPE, 5 - NLI)}
\end{figure}

\begin{figure}[t]
\centering
\begin{subfigure}{0.5\textwidth}
    \includegraphics[width=\textwidth]{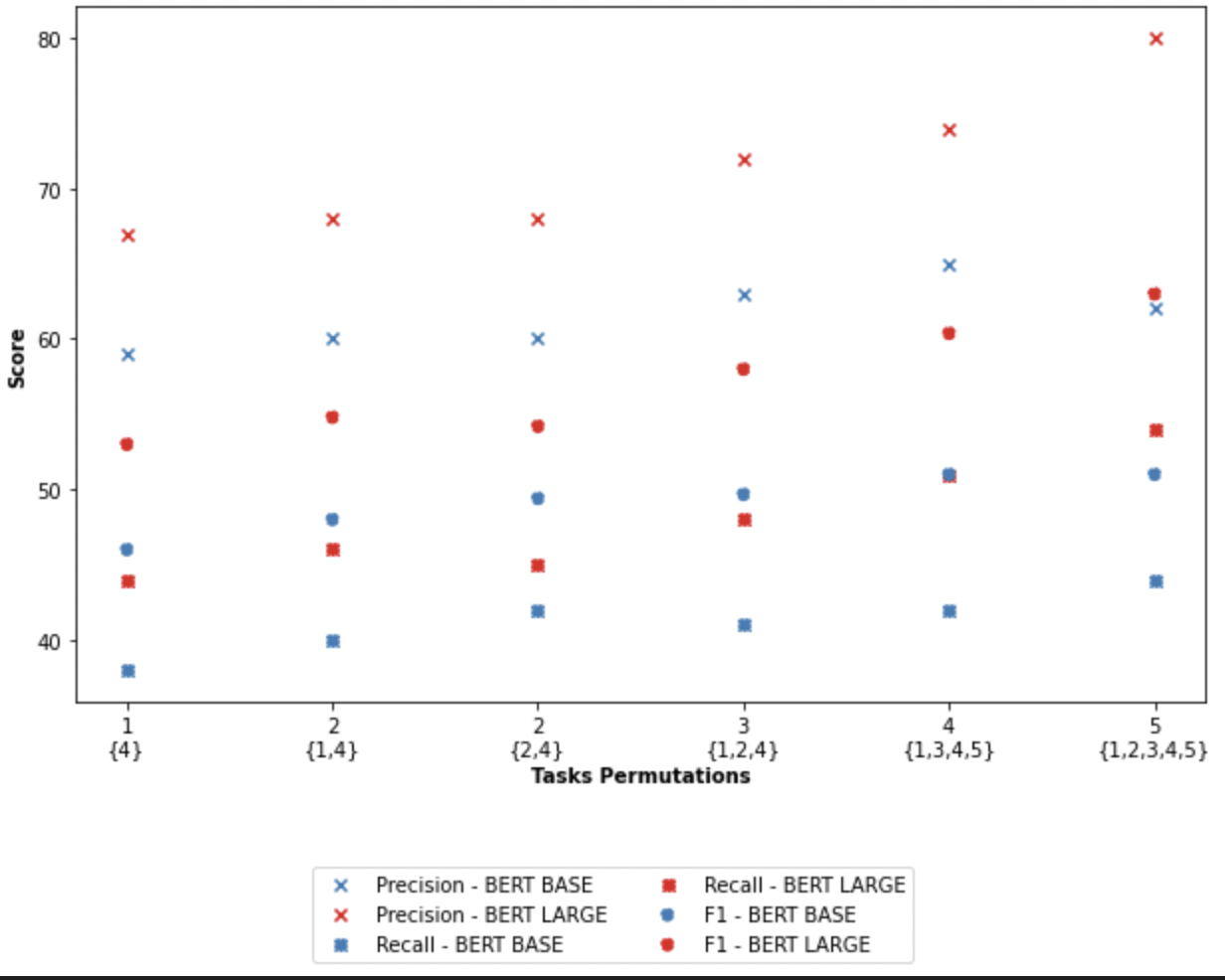}
    \caption{NPE}
    \label{Fig:CoreferenceRes}
\end{subfigure}
\hfill
\begin{subfigure}{0.5\textwidth}
    \includegraphics[width=\textwidth]{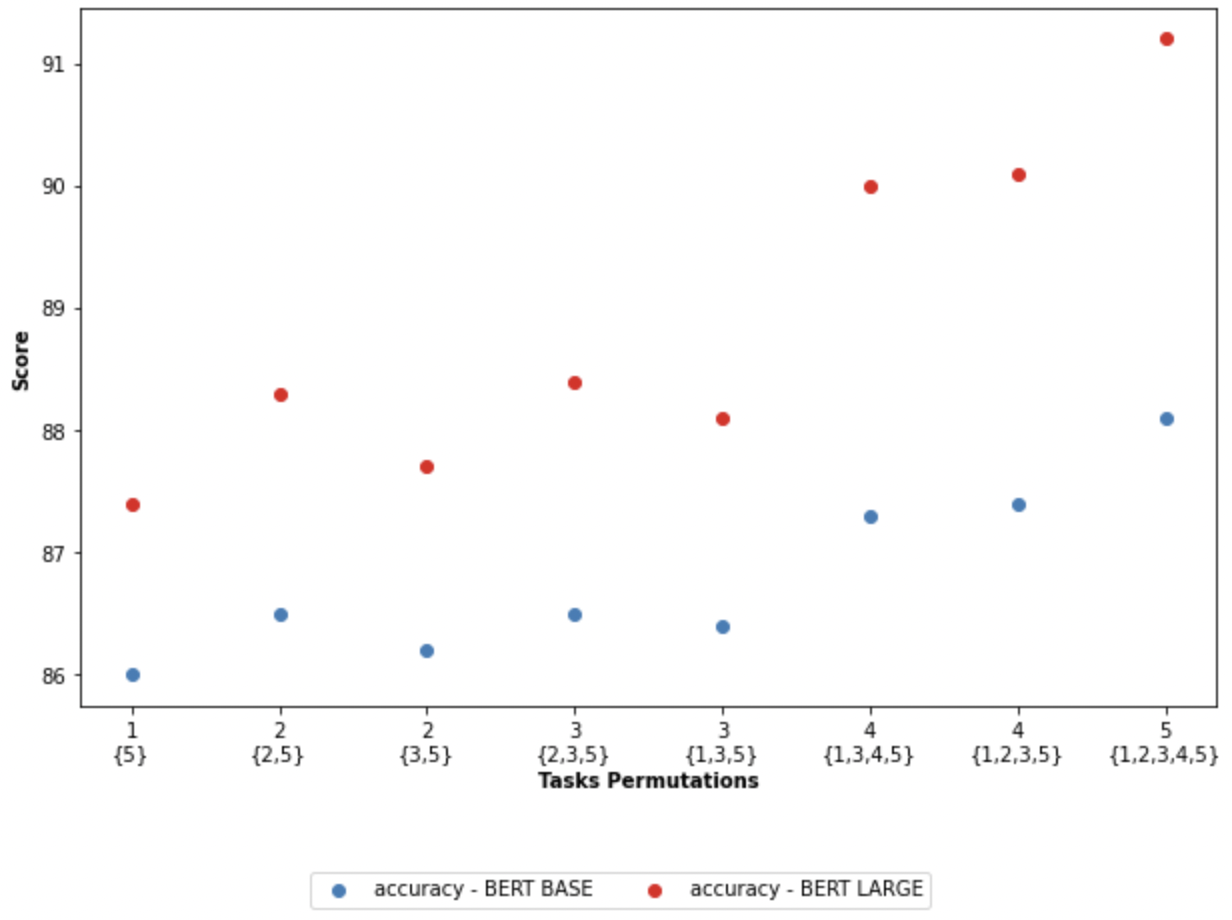}
    \caption{NLI}
    \label{Fig:NLIRes}
\end{subfigure}
\hfill
\caption{Results for all tasks, for different permutations of tasks fine-tuned upon. The labels are the number of tasks and in curly brackets which tasks (1 - SRO, 2 - ISR, 3 - DRR, 4 - NPE, 5 - NLI)}
\end{figure}

Figures~\ref{Fig:ReorderRes}, \ref{Fig:RelevanceRes}, \ref{Fig:DiscourseRes}, \ref{Fig:CoreferenceRes} and \ref{Fig:NLIRes} visualize the performance of coherence proxy tasks across fine-tuning settings for BERT-base and BERT-large models. It highlights how subsets of tasks impacts target task performance.

\section{T5 Prompts and Outputs for Different Tasks} \label{App:T5Prompts}
In Table~\ref{Tab:PromptsT5} we detail the various prompts used for fine-tuning T5 models on all explored tasks in this work.
\begin{table*}[t]
    \centering
    \begin{tabularx}{\textwidth}{|l|l||X|}
    \hline
    \hline
         Task Name         & Dataset Name & Prompt       \\
         \hline
         \hline
         SRO               & ROCStories   &``reorder: what is the order of the sentences so that the paragraph is coherent? sentence 1: $\langle S_1 \rangle$ sentence 2: $\langle S_2 \rangle$ ... $\langle S_N\rangle$'' \\
         \hline
         ISR               & ROCStories   &``relevance: what is the irrelevant sentence in the text? sentence1: $\langle S_1 \rangle$ sentence2: $\langle S_2 \rangle$ sentence3: ...$\langle S_N \rangle$'' \\
         \hline
         DRR               & PDTB3        &``discourse relation: what is the discourse relation between $\langle DU_1 \rangle \langle DU_2 \rangle$'' \\
         \hline
         NPE               & TNE          &``coreference text: what are the preposition relations between <$NP_i$> and <$NP_j$>? text: <$P$>'' \\
         \hline
         NLI               & MNLI         &``mnli: does this hypothesis contradict, entail, or neutral with the premise? hypothesis: $\langle H \rangle$ premise: $\langle P \rangle$'' \\ 
         \hline
         \hline
         Coherence Scoring & GCDC         & ``GCDC coherence: what is the coherence score of the text (3 - high, 1 - low)? text: $\langle P \rangle$'' \\ 
         \hline
         Coherence Scoring & CoheSentia   & ``CoheSentia coherence: what is the coherence score of the text (5 - high, 1 - low)? title: $\langle T \rangle$ text: $\langle P \rangle$'' \\ 
         \hline
         \hline
         MT                & WMT14       & ``Machine Translation: what is the translation of the next text from language $<source\_language>$ to $<target\_language>$?: text in source language'' \\ 
         \hline
         NER               & Conll2003   & ``NER task: what is the entity recognition tagging of each token in the next text? $<extra\_id\_0>$ token1 $<extra\_id\_1>$ token2 ...'' \\ 
         \hline
         POS               & Conll2003   & ``POS task: What is the part of speech tagging of each token in the next text? $<extra\_id\_0>$ token1 $<extra\_id\_1>$ token2 ...'' \\ 
         \hline
         \hline
         Cohesion Reasoning & CoheSentia & ``Cohesion reasoning: previous data: <$d_i$> new sentence: <$si$>. Task: is the new sentence cohesive in regard to the previous data? give a yes or no answer to each item ''     \\
         \hline
         Consistency Reasoning & CoheSentia & ``Consistency reasoning: previous data: <$d_i$> new sentence: <$si$>. Task: is the new sentence consistent in regard to the previous data? give a yes or no answer to each item ''     \\
         \hline
         Relevance Reasoning & CoheSentia & ``Relevance reasoning: previous data: <$d_i$> new sentence: <$si$>. Task: is the new sentence relevant in regard to the previous data? give a yes or no answer to each item ''     \\
         \hline
         \hline
    \end{tabularx}
    \caption{Prompts for all tasks in this paper when using T5 model as the backbone model}
    \label{Tab:PromptsT5}
\end{table*}

In Table~\ref{Tab:TargetsT5} we detail the various outputs used for fine-tuning T5 models on all explored tasks in this work.
\begin{table*}[t]
    \centering
    \begin{tabularx}{\textwidth}{|m{9em}|m{5em}||X|}
        \hline
        \hline
         Task              & Dataset      &  Outputs       \\
         \hline
         \hline
         SRO               & ROCStories   & list of position markers $[Y_1, Y_2, ..., Y_N]$ ($Y_i$-position of the $i_{th}$ sentence of the corresponding ordered sequence $S_i$ in the shuffled input) \\
         \hline
         ISR               & ROCStories   &the index of the irrelevant sentence in the paragraph \\
         \hline
         DRR               & PDTB3        &``$\langle$connector$\rangle \rightarrow \langle$$l_1$ relation$\rangle \rightarrow \langle$$l_2$$\rangle$'' \\
         \hline
         NPE               & TNE          &the preposition \\
         \hline
         NLI               & MNLI         &Contradict / Entails / Neutral  \\ 
         \hline 
         \hline
         Coherence scoring & GCDC         & the score \\ 
         \hline
         Coherence scoring & CoheSentia   & the score \\ 
         \hline
         \hline
         MT                & WMT14       & the translated text \\ 
         \hline
         NER               & Conll2003   & ``$<extra\_id\_0>$ $ner\_tag\_token1$ $<extra\_id\_2>$ $ner\_tag\_token2$ ...'' \\
         \hline
         POS               & Conll2003   & ``$<extra\_id\_0>$ $pos\_tag\_token1$ $<extra\_id\_2>$ $pos\_tag\_token2$ ...'' \\ 
         \hline
         \hline
         Cohesion reasoning & CoheSentia &  Yes / No \\
         \hline
         Consistency reasoning & CoheSentia &  Yes / No \\
         \hline
         Relevance reasoning & CoheSentia &  Yes / No \\
         \hline
         \hline
    \end{tabularx}
    \caption{Outputs for all tasks in this paper when using T5 model as the backbone model}
    \label{Tab:TargetsT5}
\end{table*}





\end{document}